\documentclass{elsarticle}
\usepackage{fullpage}
\makeatletter
\def\ps@pprintTitle{%
 \let\@oddhead\@empty
 \let\@evenhead\@empty
 \def\@oddfoot{\reset@font\hfil\thepage\hfil}
 \let\@evenfoot\@oddfoot
 }
 \makeatother

\usepackage{amsmath}
\usepackage{amssymb}
\usepackage{amsfonts}
\usepackage{amssymb}
\usepackage{graphicx}

\usepackage{color}
\usepackage{bm}
\usepackage{multirow}
\usepackage{booktabs}
\usepackage{soul}
\usepackage[raggedright]{sidecap}
\usepackage{subcaption}

\usepackage{ulem}
\usepackage{cancel}
\def\dc{DDC}
\def\x{{\mathbf x}}

\DeclareMathOperator{\triu}{triu}

\biboptions{sort&compress}

\usepackage{algorithm}
\usepackage{algorithmic}

\usepackage{url}

\usepackage{subcaption}

\begin{document}

\title{Deep Divergence-Based Approach to Clustering}

\author[1]{Michael Kampffmeyer}
\author[1]{Sigurd L{\o}kse}
\author[1]{Filippo M. Bianchi}
\author[2,3]{Lorenzo Livi}
\author[4]{Arnt-B{\o}rre~Salberg}
\author[1,4]{Robert Jenssen}
\address[1]{Machine Learning Group, UiT the Arctic University of Norway, http://machine-learning.uit.no/}
\address[2]{Department of Computer Science, University of Exeter, UK}
\address[3]{Departments of Computer Science and Mathematics, University of Manitoba, Canada}
\address[4]{Norwegian Computing Center, Oslo, Norway}

\begin{abstract}
A promising direction in deep learning research consists in learning representations and simultaneously discovering cluster structure in unlabeled data by optimizing a discriminative loss function.
As opposed to supervised deep learning, this line of research is in its infancy, and how to design and optimize suitable loss functions to train deep neural networks for clustering is still an open question.
Our contribution to this emerging field is a new deep clustering network that leverages the discriminative power of information-theoretic divergence measures, which have been shown to be effective in traditional clustering. 
We propose a novel loss function that incorporates geometric regularization constraints, thus avoiding degenerate structures of the resulting clustering partition.
Experiments on synthetic benchmarks and real datasets show that the proposed network achieves competitive performance with respect to other state-of-the-art methods, scales well to large datasets, and does not require pre-training steps.
\end{abstract}

\maketitle
\section{Introduction}
Deep neural networks~\cite{krizhevsky2012imagenet,Goodfellow-et-al-2016} excel at hierarchical representation learning~\cite{bengio2013representation}, and yield state-of-the-art performance in image classification~\cite{krizhevsky2012imagenet}, object detection~\cite{renNIPS15fasterrcnn}, segmentation~\cite{long2015fully,kampffmeyer2016semantic}, time series prediction~\cite{bianchi2017overview} and speech recognition~\cite{graves2013speech}, to name a few.
However, deep networks are usually trained in a supervised manner, hence requiring a large amount of labeled data.
This is a challenge in many application domains.

Clustering \cite{jain2010data,von2007tutorial}, one of the fundamental areas in machine learning, aims at categorizing unlabeled data into groups (clusters). A promising direction in deep learning research is to learn representations and simultaneously discover cluster structure in unlabeled data by optimizing a discriminative loss function.
Deep Embedded Clustering (DEC)~\cite{xie2015unsupervised} exemplifies this line of work and represents, to the best of our knowledge, the state-of-the-art.
DEC is based on an optimization strategy in which a neural network is pre-trained by means of an autoencoder and then fine-tuned by jointly optimizing cluster centroids in output space and the underlying feature representation. Another example is~\cite{yang2016towards}, where the authors propose a joint optimization for dimensionality reduction using a neural network and $k$-means clustering.
Alternative approaches to unsupervised deep learning based on adversarial networks have recently been proposed~\cite{goodfellow2014generative}. These approaches are different in spirit but can also be used for clustering~\cite{springenberg2015unsupervised, makhzani2015adversarial}.

\begin{figure}[tbp]
\centering
\includegraphics[width=0.6\columnwidth, keepaspectratio, trim={0cm 0cm 1cm 0cm},clip]{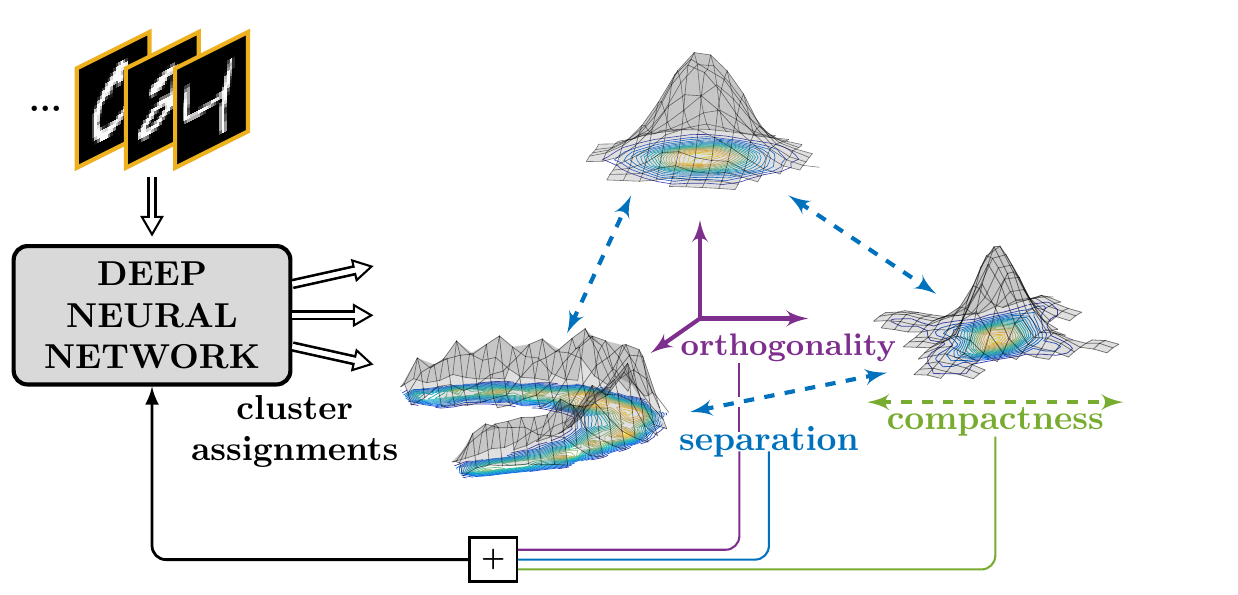}
\caption{Our approach takes advantage of the power of deep learning to extract features and perform clustering in an end-to-end manner. The proposed loss function is rooted in two fundamental objectives of clustering: separation and compactness of clusters.}
\label{fig:motivation}
\end{figure}

In this work, we propose what we called the Deep Divergence-based Clustering (DDC) algorithm.
Our method takes inspiration from the vast literature on traditional clustering techniques that optimize discriminative loss functions based on information-theoretic measures~\cite{dhillon2003divisive,jenssen2007information,tishby2001data,Vikjord20143070}.
The main motivation for this choice is that the divergence, as a measure of dissimilarity between clusters represented by their probability density functions, builds on two fundamental objectives (Figure \ref{fig:motivation}): the separation between clusters and the compactness within clusters. These are desirable properties to increase identifiability of nonparametric mixtures~\cite{aragam2018identifiability}.
Our new divergence-based loss function for deep clustering supports end-to-end learning and explicitly exploits knowledge about the geometry of the output space during the optimization. \dc{} achieves state-of-the-art performance without requiring hand-crafted feature design, reducing also the importance of a pre-training phase.

A preliminary version of this paper appeared in~\cite{kampffmeyer2017mlsp}. The preliminary version was targeted towards image clustering combining a convolutional neural network architecture with our proposed clustering loss function.
Here, we extend our work by 
(i) modifying the proposed architecture such that it can also handle textual data;
(ii) conducting experiments and comparisons on additional datasets (including textual data -- Reuters dataset);
(iii) providing a thorough analysis of the proposed cost function and its components via ablation experiments; 
(iv) illustrating and discussing the functioning of the method in controlled settings;
(v) interpreting predictions of the network by means of guided backpropagation~\cite{springenberg2014striving};
and finally (vi) providing a more thorough literature background discussion, placing our work into a broader context.

This paper is structured as follows. Section~\ref{sec:related} provides an overview of related works. Section~\ref{sec:method} presents the proposed methodology for performing clustering with deep networks. In Section~\ref{sec:exp}, we show the experimental results on several datasets and analyze the proposed cost function in detail. Finally, in Section~\ref{sec:conclusion} we draw conclusions and point to future directions.

\section{Related work}
\label{sec:related}
Common approaches to unsupervised deep learning include methods based on deep belief networks, autoencoders, and generative adversarial networks~\cite{bengio2013representation,goodfellow2014generative}. These methods have been mainly used for unsupervised pre-training~\cite{erhan2010does}. Deep belief networks were the first of these models to be proposed and consist of stacked restricted Boltzmann machines that are trained in a greedy fashion~\cite{hinton2006fast}. Once trained, deep belief networks can be used to initialize neural networks. 

Although several types of autoencoders have been proposed, all share a common underlying architecture consisting of an encoding and a decoding layer.
The encoder is responsible for producing a hidden representation; the decoder re-generates inputs from the hidden representation.
Both can efficiently be learned using backpropagation, by minimizing the reconstruction loss between original input and decoder output.
Variations include, among others, denoising autoencoders~\cite{vincent2008extracting}, which regularize the original autoencoder model by adding noise to inputs and then changing the objective to both include reconstruction and denoising, contractive autoencoders~\cite{rifai2011contractive}, and more recently autoencoders that are regularized by preserving similarities in input space~\cite{Kampffmeyer2017}.
Variational autoencoders~\cite{kingma2013auto} have been used recently for several unsupervised tasks, such as image generation~\cite{gregor2015draw} and segmentation~\cite{sohn2015learning}. This approach assumes that data are generated from directed graphical models and uses a variational approach to learn latent representations.

Adversarial generative models~\cite{goodfellow2014generative} are more recent approaches to unsupervised deep learning. Here, two networks are trained: one is responsible for discriminating between real and generated images; the other is responsible for generating realistic-enough images to confuse the first network. 

Clustering is a classic information processing problem, particularly important in machine learning~\cite{jain2010data,rodriguez2014clustering,cagata,nie2018multiview,myhre2018robust}. Countless approaches exist for clustering, with mean shift~\cite{comaniciu2002mean}, $k$-means and expectation--maximization algorithms~\cite{aggarwal2013data}, being some of the most well-known ones.
In the last decade, \textit{spectral clustering} played a prominent role in the field, see for instance~\cite{jenssenKECA,ng2002spectral,nie2011spectral,von2007tutorial,yang2010image}.
Spectral clustering exploits the spectrum of similarity matrices to partition input data.
Although these methods have demonstrated good performance in complex problems, they suffer from lack of scalability with respect to the number of input data points; cubic computational complexity for eigensolvers and quadratic complexity in terms of memory occupation.
Attempts to solve these problems have been made by designing approximations or employing different optimization techniques \cite{dhillon2004kernel,han2016mini,yan2009fast}.

Only a few methods have been proposed to exploit deep learning architectures for clustering, thereby taking advantage of hierarchical feature representations learned by such networks.
CatGAN~\cite{springenberg2015unsupervised}, and AAE~\cite{makhzani2015adversarial} are based on the idea of adversarial networks.
CatGAN is a method for learning a discriminative model, trained by optimizing a loss function implementing two different objectives. The first accounts for mutual information and predicted categorical distribution of classes in the data. The second objective maximizes the robustness of the discriminative network against an adversarial generative model.
AAE instead assumes that data are generated from two latent variables, one associated with a categorical distribution and the other with a Gaussian distribution, and uses two adversarial networks to impose these distributions on the data representation.
In a recent contribution \cite{bojanowski2017unsupervised}, the authors propose an unsupervised training algorithm for CNNs and test its performance on image classification problems. The idea is to deal with the so-called ``feature collapse problem'' by mapping the learned features on random targets uniformly distributed on a $d$-dimensional unit sphere. A combination of recurrent and convolutional networks has also been used to perform image clustering by interpreting agglomerative clustering as a recurrent process~\cite{yang2016joint}. Another recent approach to clustering based on the idea of hierarchical feature representation learning is provided by Zhang~\cite{zhang2018multilayer}, who proposes a multilayer bootstrap network where each layer performs multiple mutually independent k-centroids clusterings. Each layer gets trained individually in a bottom-up fashion and the input of consecutive layers is an indicator vector of which centroids are closest to a given input. Unlike the previously discussed methods, the multilayer bootstrap network does not offer end-to-end training.

To the best of our knowledge, DEC~\cite{xie2015unsupervised} is the method that is most closely related to our approach, as it is also founded on traditional clustering approaches. DEC simultaneously learns a feature representation as well as a cluster assignment in a two-step procedure. In the first step, soft assignments are computed between the data and cluster centroids based on a Student's t-distribution. Then, the parameters are optimized by matching soft assignments to a target distribution, which expresses confidence in assignments. The matching is performed by minimizing the Kullback-Leibler divergence. 
However, the effectiveness of DEC depends on a  pre-training step implemented with autoencoders and does require explicit feature design to handle complex image data, e.g., Histogram of Oriented Gradients (HOG) features~\cite{dalal2005histograms}.

\section{Deep clustering}
\label{sec:method}
We first describe, in Section \ref{sec:loss} the proposed clustering loss function and present a description of the overall algorithm in Section~\ref{sec:descAlg}.
Successively, in Section \ref{sec:arch}, we discuss the deep network architectures that we propose to use for clustering problems, namely one that is based on convolutional layers for image clustering and one based on fully connected layers for vectorial data. 
Finally, we discuss scalability in Section~\ref{sec:memory}.

Inspired by recent successes of introducing companion losses \cite{lee2015deeply} to supervised deep learning models, we propose a loss function for clustering that includes terms computed over several network layers. 
The details of the loss function are outlined in what follows.

\begin{figure}[tbp]
\centering
\includegraphics[width=0.8\columnwidth, keepaspectratio]{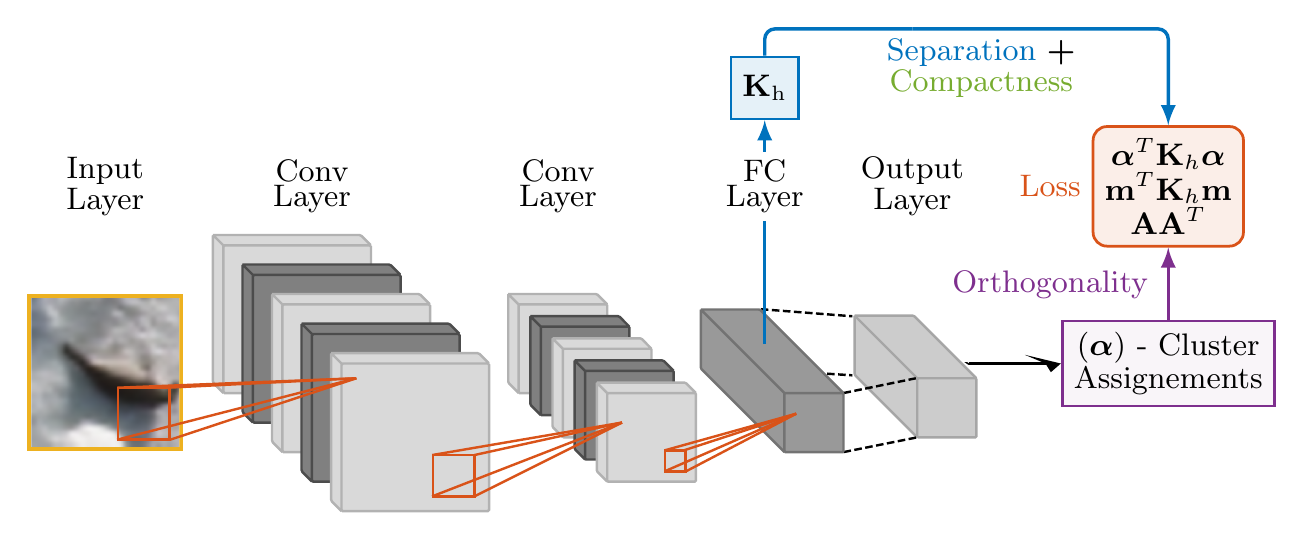}
\caption{Schematic depiction of the proposed architecture for image datasets and details of the loss function. The network consists of two convolution layers (each one followed by a pooling layer, not depicted in the figure) and a fully connected (FC) layer. Each layer is followed by a non-linear ReLU transformation. Finally, a fully connected output layer implements a logistic function (softmax). The unsupervised loss function operates on the kernel matrix $\mathbf{K}_h$ encoding data similarities evaluated on the hidden representation, and the values of the cluster assignment returned by the softmax function. The orthogonality constraint is derived from cluster assignments, while separation and compactness constraints come from the Cauchy-Schwarz divergence, computed on the similarity matrix (weighted by cluster assignments). The convolutional layers are replaced when non-image data are considered.}
\label{fig:arch}
\end{figure}

\subsection{The loss function for clustering}
\label{sec:loss}
The design of a loss function that allows the network to learn via gradient descent the intrinsic cluster structure in the input data is a fundamental part of this work. As illustrated in Figure~\ref{fig:arch} and explained below, in addition to exploiting the geometry of the output space induced by the softmax activation, we adopt a kernel-based approach to estimate the divergence between clusters.

\subsubsection{Loss term based on information-theoretic divergence measures}
An information-theoretic divergence measure computes the dissimilarity between probability density functions (PDFs).
For a clustering application, one would model each cluster by its PDF and optimize cluster assignments such that the divergence between their PDFs is maximized.
Several different formulations of divergence measures exist in the literature~\cite{Basseville2013621}, many of which are suitable for clustering.
In this work, we focus on one particular divergence measure that has been proven useful for clustering in the past, namely the Cauchy-Schwarz (CS) divergence~\cite{jenssen2006cauchy, Vikjord20143070}, also referred to as the Information Cut in a graph clustering perspective~\cite{jenssen2007information}. The CS divergence can be used in multi-cluster problems (i.e., problems with more than two clusters) by averaging the pairwise divergence over all pairs of cluster PDFs.

Considering $k\geq 2$ distinct PDFs, the CS divergence is defined as
\begin{equation}
\begin{aligned}
\label{eq:csdiv}
&D_\text{cs}(p_1,\dots, p_k) = - \log \left( \frac{1}{k} \sum \limits_{i=1}^{k-1} \sum \limits_{j > i} \frac{\int p_i(\x)p_j(\x) d\x}{\sqrt{\int p_i^2(\x)d\x \int p_j^2(\x) d\x}} \right).
\end{aligned}
\end{equation}
For a pair of PDFs, $p_i$ and $p_j$, we have $0 \leq D_\text{cs}(p_1, p_2) < \infty$, where we obtain the minimum value iff $p_i = p_j$. 
Thus, in order to maximize cluster separation and compactness, we want the divergence to be as large as possible.
Since the logarithm is a monotonic function, maximizing (\ref{eq:csdiv}) is in practice equivalent to minimizing the argument of the logarithm.
We observe that the minimum is obtained when the numerator is small and the denominator is large. 
Intuitively, this fact implies that the similarity between samples in different clusters is small (numerator) and similarity of samples within the same cluster is large (denominator).

In this paper, we make use of the divergence in (\ref{eq:csdiv}) to measure the distance between clusters.
Since the underlying true densities $p_i$, $p_j$ are unknown, we follow a data-driven approach and approximate the PDFs using a Parzen window estimator, configured with a Gaussian kernel having bandwidth $\sigma$.
We define the matrix $\mathbf{K} \in \mathbb{R}^{n \times n}$ that encodes the similarities between $n$ input data.
The matrix element $k_{i,j}$ stores the value $\exp(-d(\mathbf x_i, \mathbf x_j)^2/(2\sigma^2))$, where $d(\mathbf x_i, \mathbf x_j)$ is the Euclidean distance between data point $\mathbf x_i$ and $\mathbf x_j$.
Using the Parzen window estimator, the CS divergence can be expressed as \cite{jenssen2006cauchy}
\begin{equation}
\label{eq:csdiv_parzen}
D_\text{cs} =
- \log
    \left(
    \frac{1}{k} \sum \limits_{i=1}^{k-1} \sum \limits_{j > i}
    \frac{
        \sum \limits_{q\in C_i} \sum \limits_{l \in C_j} k_{q,l}
    }{
        \sqrt{
            \sum \limits_{q, q' \in C_i} k_{q,q'} 
            \sum \limits_{l, l' \in C_j} k_{l,l'} 
        }
    }
    \right).
\end{equation}
Note that this estimate in \eqref{eq:csdiv_parzen} of the CS divergence can also be interpreted as measuring the cosine of the angle between cluster means in a Reproducing Kernel Hilbert Space~\cite{jenssen2006cauchy} and is closely related to maximum mean discrepancy~\cite{gretton2012kernel}. 
Assume that we have a $n \times k$ cluster assignment matrix $\mathbf A = [\alpha_{q,i}]$, with elements $\alpha_{q,i} \in \{0,1\}$ that represent the crisp cluster assignment of data point $q$ to cluster $C_i$. Thus, each data point $q$ is represented by a one-hot vector. 
Then,
\begin{equation}
    \nonumber
        \sum \limits_{q\in C_i} \sum \limits_{l \in C_j} k_{q,l}
        = \sum \limits_{q, l = 1}^n \alpha_{q, i} \alpha_{l, j} k_{q,l}
        = \boldsymbol{\alpha}_{i}^T \mathbf K \boldsymbol{\alpha}_j,
\end{equation}
where $\boldsymbol \alpha_i$ is the $i$-th column of $\mathbf A$.
The CS-divergence becomes $D_\text{cs} = - \log\left(d_{\alpha}\right)$, where
\begin{align}
\label{eq:delta_sim}
d_{\alpha} &= \frac{1}{k} \sum \limits_{i=1}^{k-1} \sum \limits_{j > i}
  \frac{
    \boldsymbol \alpha_i^T \mathbf K \boldsymbol \alpha_j
  }{
    \sqrt{
        \boldsymbol \alpha_i^T \mathbf K \boldsymbol \alpha_i
        \boldsymbol \alpha_j^T \mathbf K \boldsymbol \alpha_j
    }
  }.
\end{align}

The formulation of the CS-divergence in \eqref{eq:csdiv_parzen} generalizes to soft cluster assignments, $\alpha_{q, i}$, preserving the differentiability of the loss function. In our architecture, the soft cluster assignments correspond to the softmax outputs and thereby the probability of a data point $q$ to belong to cluster $C_i$.

The similarity values in $\mathbf{K}$ depend on the data representation. In particular, as data are processed by the neural network, several non-linear transformations map inputs onto different feature spaces, representing different levels of abstraction. The kernel bandwidth $\sigma$ is computed based on the statistics of the learned representations. More details can be found in Section \ref{sec:Implementation}.
To take advantage of the different representations and use the idea of companion losses for restricting the intermediate representations of the network, we use the hidden representation computed by the last fully connected layer before the output layer in addition to the soft cluster assignments produced by the softmax output layer. 
We do this by computing a (kernel) similarity matrix $\mathbf{K}_\text{hid}$, which, by considering the corresponding $d_\text{hid,$\alpha$}$ in \eqref{eq:delta_sim}, yields a similarity score.

\subsubsection{Loss term based on the geometry of the output space}
\label{sec:regularization}

The output space has a fixed number of dimensions (corresponding to the number of output neurons/clusters) and a precise geometry induced by the softmax activations used in the output layer (whose elements sum to 1), which we exploit in our algorithm:
\begin{enumerate}
    \item The output space is a simplex in $\mathbb R^k$;
    \item A data points degree of membership to a given cluster is maximized if the cluster assignment lies in a corner of the simplex (i.e., $\alpha_{q, i} = 1$ if data point $q$ is fully assigned to cluster $C_i$);
    \item Following from the previous point, cluster assignment vectors of data points assigned to different clusters, in the optimal case, should be orthogonal to each other.
\end{enumerate}

This intuition about the geometry enables us to introduce a term that avoids degenerate solutions by addressing the aforementioned problem of collapsing features/clusters and encourages diversity in cluster assignment.
For a given cluster assignment matrix, $\mathbf A$, the strictly upper triangular elements of $\mathbf A \mathbf A^T$, denoted by $\triu(\mathbf A \mathbf A^T)$, consists of inner products between cluster assignment vectors. Unless explicitly stated, $\triu(\mathbf A \mathbf A^T)$ will denote the sum of these elements. 
Note that we do not include the elements on the diagonal. Further, $\mathbf A \mathbf A^T$ will consist entirely of non-negative elements because $\mathbf A$ is non-negative; cluster assignment vectors are orthogonal if and only if these inner products are zero.
Thus, our criterion consists of enforcing low values in the upper triangular elements.
This also has the effect of a regularization term if the number of clusters is smaller than the number of input data points. Indeed, not all data points in the restricted space can be orthogonal to each other, forcing data points to repel each other, thereby acting against the terms that try to improve similarity. This term also encourages a balanced distribution of data points in the different classes, which makes our loss ideal for problems with balanced classes. Alternative regularization methods that are not based on the balanced class assumption will be investigated in future work.

The fact that cluster assignment vectors are orthogonal, however, does not imply that such vectors are embedded in a corner of the simplex.
As an example, assume that $\alpha_{q, i} = 1$ and $\alpha_{l, i} = 0$.
Due to the restrictions of the simplex geometry, it follows that $\alpha_{q,k}=0,\; k\ne i$ and therefore $\boldsymbol\alpha_{q}^{T}\boldsymbol\alpha_{l} = 0$ independently of the values of $\alpha_{l,k},\;k\ne i$. Thus, $l$ is not restricted to a simplex corner. 
Therefore, in order to enforce closeness to a corner of the simplex, we define an additional term for the loss function that reads
\begin{equation*}
    m_{q, i} = \exp(-\|\boldsymbol\alpha_q - \mathbf e_i\|^2),
\end{equation*}
where $\mathbf e_i \in \mathbb{R}^k$ is a vector denoting the $i$th corner of the simplex; representing cluster $C_i$.
This exponential evaluates to one only when $\boldsymbol\alpha_q$ is located in a corner of the simplex.
We make use of this fact by defining a third similarity term $d_\text{hid,m}$, where $\mathbf m_i = [m_{q,i}] \in \mathbb{R}^n$ takes the place of $\boldsymbol \alpha_i$ in \eqref{eq:delta_sim}.

\subsubsection{The final clustering loss function}
The weights in the neural network architecture described in Section~\ref{sec:arch} can then be trained by minimizing the sum of the three different terms discussed in the previous section:
\begin{equation}
\begin{aligned}
    \label{eq:loss_fun}
    L &= d_\text{hid,$\alpha$} + \triu(\mathbf A \mathbf A^T) + d_\text{hid,m} \\
      &= \frac{1}{k} \sum \limits_{i=1}^{k-1} \sum \limits_{j > i}
  \frac{
    \boldsymbol \alpha_i^T \mathbf K_{hid} \boldsymbol \alpha_j 
  }{
    \sqrt{
        \boldsymbol \alpha_i^T \mathbf K_{hid} \boldsymbol \alpha_i
        \boldsymbol \alpha_j^T \mathbf K_{hid} \boldsymbol \alpha_j
    }
  } + \triu(\mathbf A \mathbf A^T) + \frac{1}{k} \sum \limits_{i=1}^{k-1} \sum \limits_{j > i}
  \frac{
    \mathbf{m}_i^T \mathbf K_{hid} \mathbf m_j 
  }{
    \sqrt{
        \mathbf m_i^T \mathbf K_{hid} \mathbf m_i
        \mathbf m_j^T \mathbf K_{hid} \mathbf m_j
    }
  }
\end{aligned}
\end{equation}
\begin{figure*}[tbp!]
\centering
\begin{subfigure}[b]{0.30\columnwidth}
  \centering
  \includegraphics[width=\textwidth, keepaspectratio,clip]{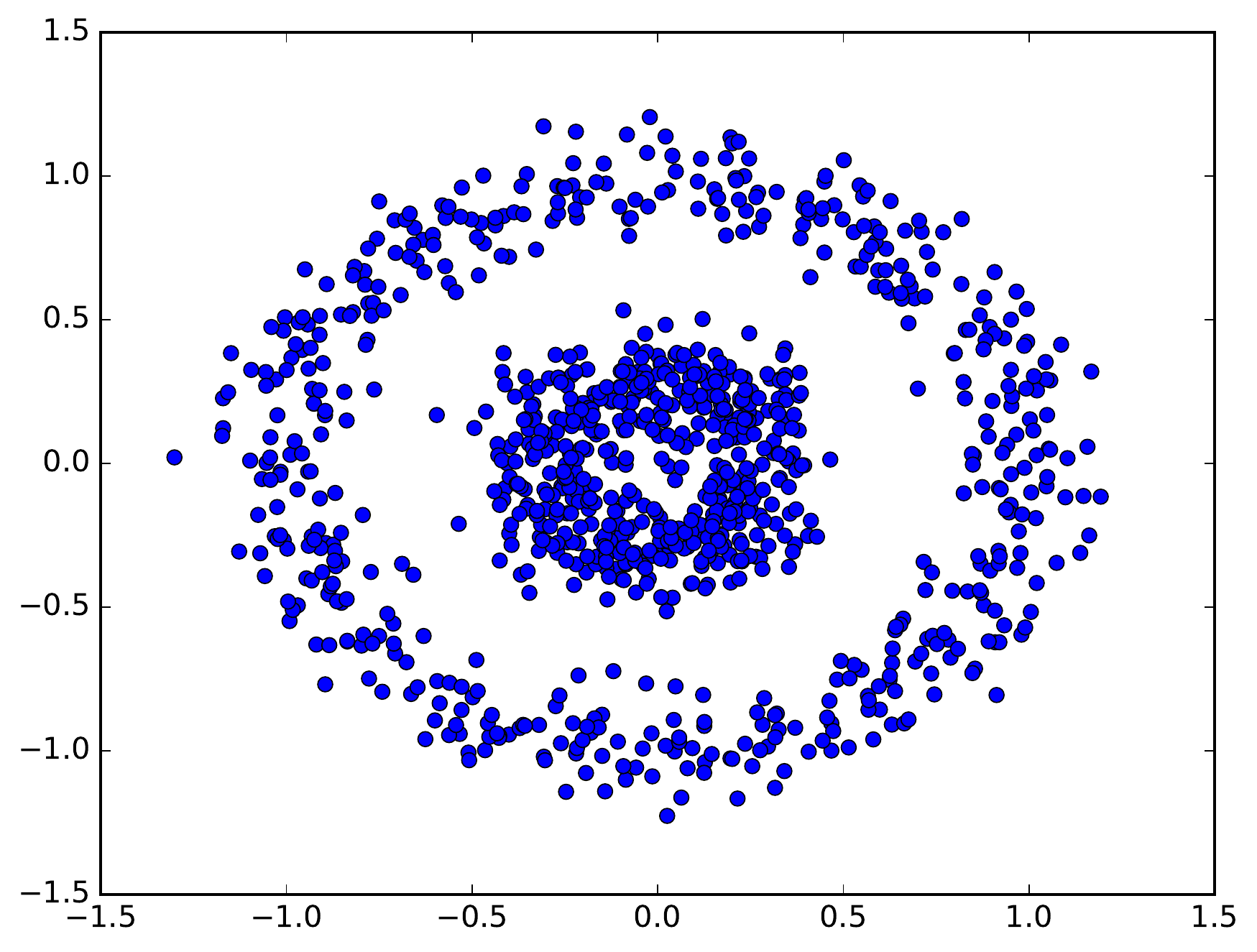}
  \caption{Sample dataset.}
  \label{fig:circles_data}
\end{subfigure}%
\quad
\begin{subfigure}[b]{0.30\columnwidth}
  \centering
  \includegraphics[width=\textwidth, keepaspectratio,clip]{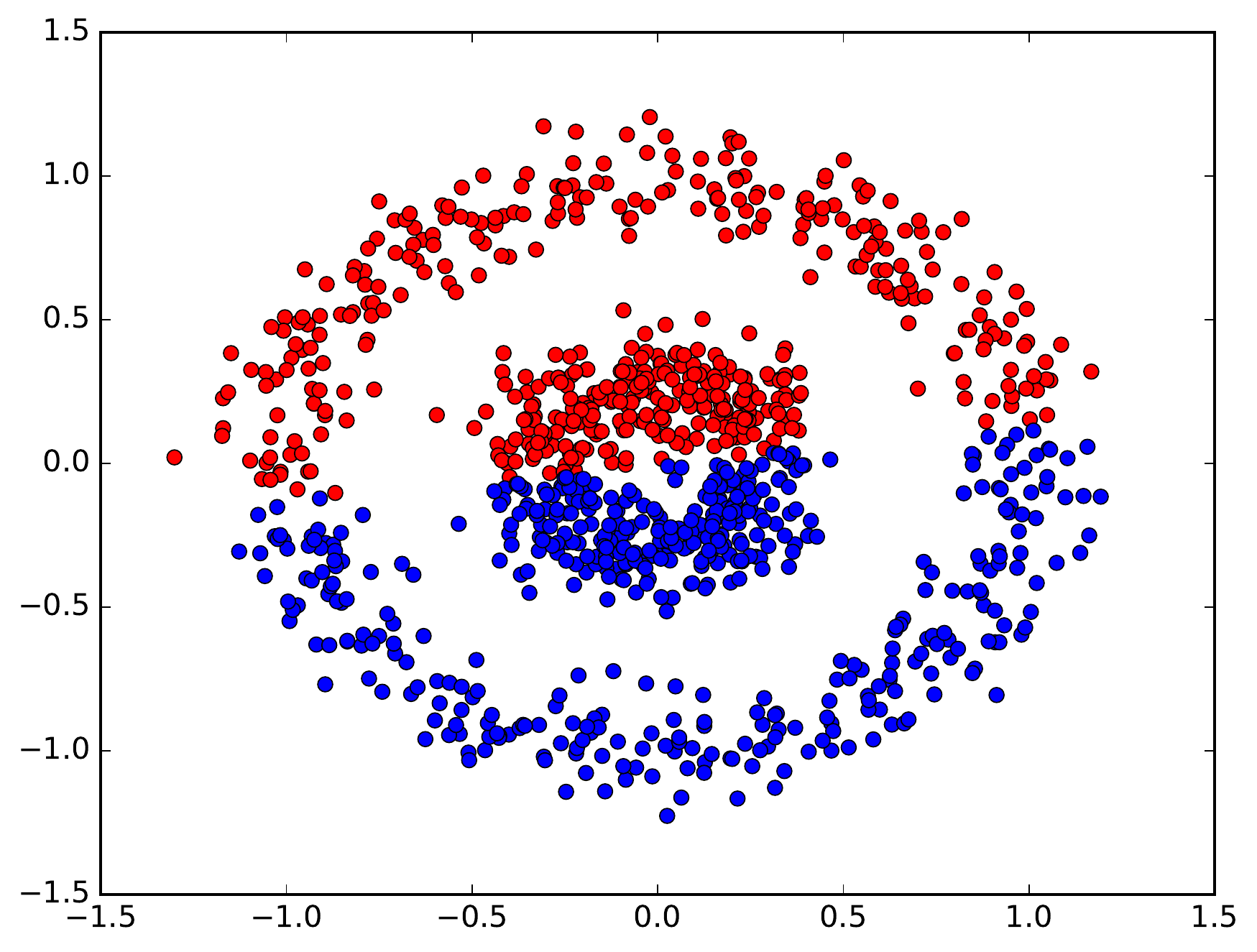}
  \caption{Outcome of $k$-means.}
  \label{fig:circles_kmeans}
\end{subfigure}
\quad
\begin{subfigure}[b]{0.30\columnwidth}
  \centering
  \includegraphics[width=\textwidth, keepaspectratio,clip]{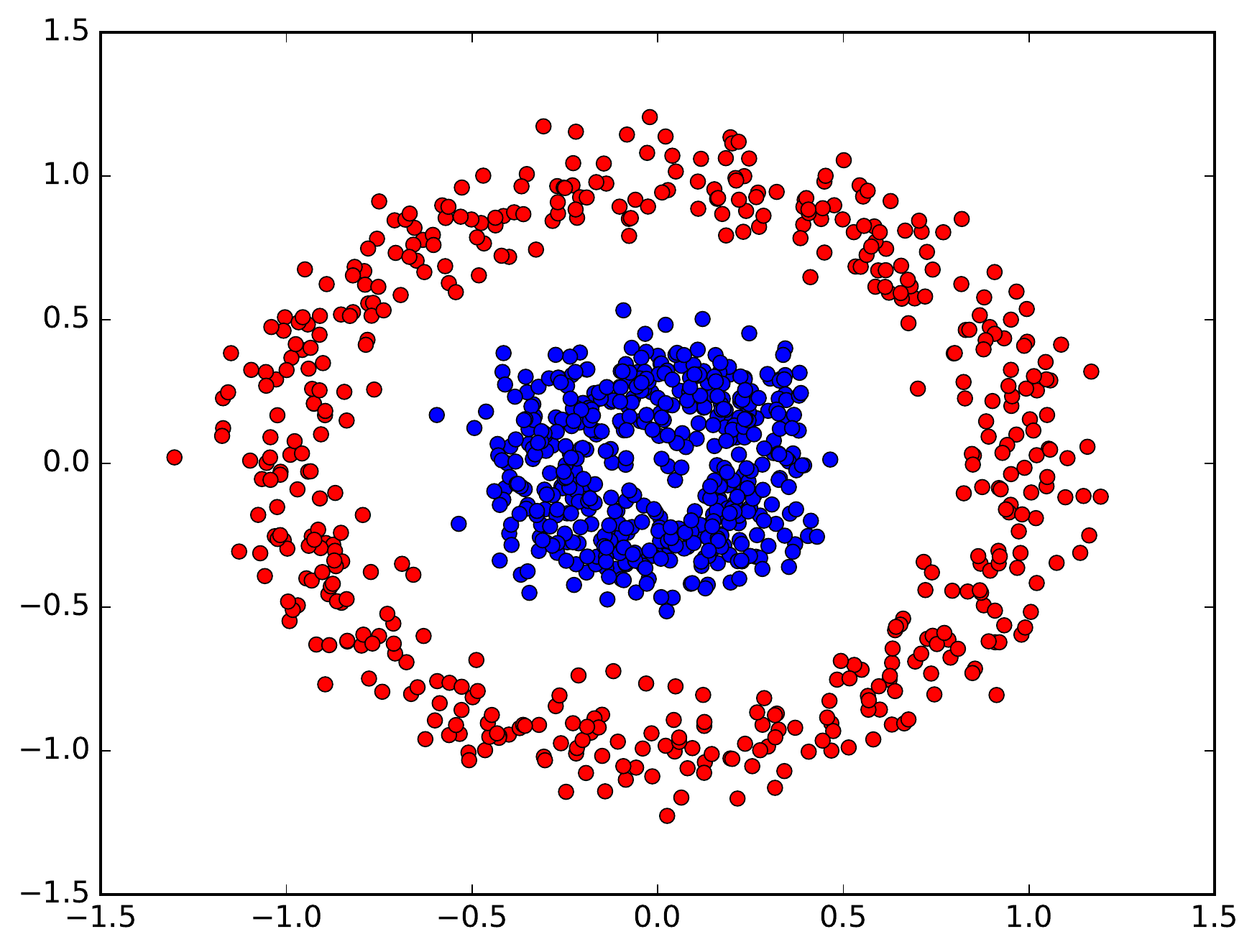}
  \caption{Outcome of \dc{}.}
  \label{fig:circles_dc}
\end{subfigure}
\caption{Illustration of \dc{} clustering outcome on a synthetic dataset, showing the capability of learning non-linear structures.}
\end{figure*}
\begin{figure*}[tbp!]
\centering
\begin{subfigure}[b]{0.30\columnwidth}
  \centering
  \includegraphics[width=\textwidth, keepaspectratio,clip]{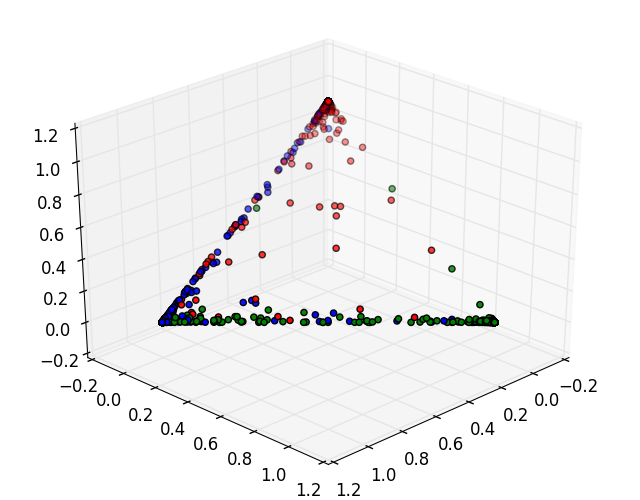}
  \caption{Epoch 0.}
  \label{fig:scatter_0}
\end{subfigure}%
\quad
\begin{subfigure}[b]{0.30\columnwidth}
  \centering
  \includegraphics[width=\textwidth, keepaspectratio,clip]{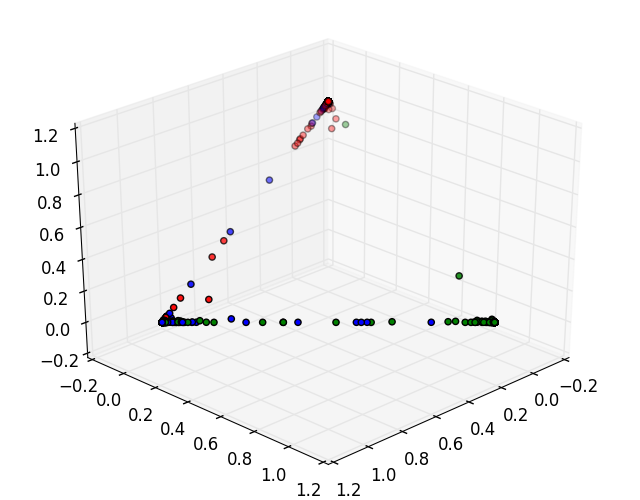}
  \caption{Epoch 5.}
  \label{fig:scatter_5}
\end{subfigure}
\quad
\begin{subfigure}[b]{0.30\columnwidth}
  \centering
  \includegraphics[width=\textwidth, keepaspectratio,clip]{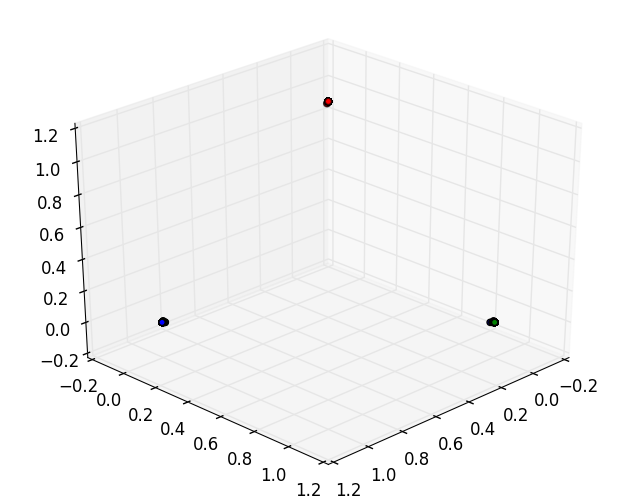}
  \caption{Epoch 15.}
  \label{fig:scatter_15}
\end{subfigure}
\caption{Illustration of \dc{} output space for three class MNIST training. Colors indicate class information of data points.}
\label{sec:scatter}
\end{figure*}

As a proof of concept, we illustrate the functioning of our clustering method on a classical synthetic dataset where one class is represented by a small circle and the other class is a ring. Note, that we use a fully-connected architecture (described in Section~\ref{sec:arch}) for these experiments as we are considering non-image data.
The dataset is shown in Figure~\ref{fig:circles_data}. Figure~\ref{fig:circles_kmeans} and \ref{fig:circles_dc} illustrate the clustering outcome using $k$-means and the proposed method, respectively. It can be observed that the proposed method captures the highly nonlinear structure in the data and is able to discover clusters of non-spherical shapes, a highly desirable quality of clustering algorithms \cite{rodriguez2014clustering}.

Further, in Figure \ref{sec:scatter} we visualize the output space for a three-cluster experiment. We chose three classes of the MNIST dataset (for dataset details see Section~\ref{sec:exp}) and visualize the output space configuration during three different stages of the optimization process. As expected, the proposed clustering loss function \eqref{eq:loss_fun} attempts to separate the data points by grouping similar points in ideal cluster centers located at the corners of the simplex. Note, we are using the convolutional architecture described in Section~\ref{sec:arch} as we are considering images in this experiment.

\subsection{Description of complete algorithm}
\label{sec:descAlg}
\begin{algorithm} 
\caption{Deep Divergence-Based Clustering}
\label{alg1}
\begin{algorithmic}[1]
    \REQUIRE $\mathcal{X}=\{\mathbf x_i\}_{i=1}^n$
    \ENSURE Cluster assignments $\mathcal{A} = \{\boldsymbol \alpha_i\}_{i=1}^n$
    \STATE Randomly initialize network parameters $\boldsymbol\Theta$
    \WHILE{not converged}
        \STATE Sample data batch $\mathcal{X}^{(b)}$ from $\mathcal{X}$
        \STATE Obtain assignment vectors $\mathcal{A}^{(b)}$ and hidden representations $\mathcal{H}^{(b)}$
        \STATE Compute $m_{q, i} = \exp(-\|\boldsymbol\alpha_q - \mathbf e_i\|^2),\,$ $\forall\; \boldsymbol\alpha_q \in \mathcal{A}^{(b)}$ 
        \STATE Estimate kernel bandwidth $\sigma$ and compute $\mathbf{K}_\text{hid}^{(b)}$ from $\mathcal{H}^{(b)}$
        \STATE Compute loss with \eqref{eq:loss_fun} and update $\boldsymbol\Theta$ with gradient descent
    \ENDWHILE
\end{algorithmic}
\end{algorithm}
In this section, we summarize the proposed algorithm. For a given data batch, the assignment vectors and the hidden representation are obtained via a single forward pass. Based on the hidden representation, the kernel bandwidth, $\sigma$, is estimated and the kernel matrix is computed. From the assignment vectors, the distance to each of the ideal cluster centers (simplex corners) is computed, obtaining $\boldsymbol{m}_q$ for each data point $q$ in a given minibatch. Using Equation~\ref{eq:loss_fun}, we then can compute the loss based on the kernel matrix, the assignment vector $\boldsymbol{\alpha}_q$, and $\boldsymbol{m}_q$. Finally, all weights in the network are updated using Adam~\cite{kingma2014adam}.
The full algorithm is outlined in Algorithm~\ref{alg1}.

\subsection{Architecture overview}
\label{sec:arch}
The network architecture is a design choice, and as such there are many options. In this paper, we choose an approach based on convolutional neural networks 
to process different image datasets, using a LeNet-inspired architecture~\cite{lecun1998gradient}. 
We selected LeNet since it is a well-known benchmark network that supports end-to-end learning and has been widely used for image classification. The architecture is depicted in Fig. \ref{fig:arch}. It consists of two convolutional layers: the first one with $32$ and the second one with $64$ $5\times5$ filters, each of them followed by a $2\times 2$ max pooling layer and a ReLU activation. 
The last convolutional layer is followed by a fully connected layer with $100$ nodes, another ReLU nonlinearity and, finally, the softmax output layer, whose dimensionality corresponds to the number of desired clusters. Batch-normalization~\cite{ioffe2015batch} is applied in the fully connected layer.
This design choice was made to increase the models' robustness and is explained in Section~\ref{sec:Implementation}.

Our approach can also be applied to cluster data that are not images simply by replacing the convolutional and pooling layers with fully connected layers. 
In particular, for the experiments conducted on non-image data, we use an architecture with four fully connected layers of size $200-200-500-C$.
The $500$ unit fully connected layer includes batch-normalization and the $C$ unit layer corresponds to the softmax output layer with dimensionality equal to the number of clusters.

Recently, theoretical advances in the theory of neural networks~\cite{giryes2016deep} highlighted how the metric structure of input data is preserved by deep neural networks with random i.i.d Gaussian weights.
One restriction is the fact that this is only true in the case where the intrinsic dimensionality of the data is proportional to the network width.
However, \cite{giryes2016deep} proved that the intrinsic dimensionality of the data does not increase as the data propagate through the network, which suggests that the width of the network (the number of hidden units per layer) that we consider for our experiments should suffice.
This theoretical property supports the design choice behind the proposed loss function, which estimates the divergence over the hidden representation, rather than in input space.

\subsection{Main memory footprint}
\label{sec:memory}
Using gradient-based optimization in neural networks allows us to process large datasets, overcoming well-known limitations of spectral methods mentioned in the introduction with regards to memory requirements.
The memory cost of our approach is kept low by the use of mini-batch training and scales linearly with the number of input data points, $n$, compared to the quadratic or super quadratic complexities encountered in spectral methods.
The proposed method scales quadratically with the mini-batch size $m$ as the kernel matrix is computed over the hidden representation for a given mini-batch; however, this is generally not an issue as $m\ll n$.

\section{Experiments}
\label{sec:exp}
\setlength{\belowcaptionskip}{0pt}
We evaluate \dc{} on the MNIST handwritten image data\-set as it represents a well-known benchmark dataset in the literature.
In addition, we evaluate our algorithm on two more challenging real-world datasets: one dataset for detection of seal pups in aerial images here referred to as the SEALS-dataset and the Reuters dataset for news story clustering. 
In the results, we compare our method to four alternative clustering approaches.

\subsection{MNIST dataset}
The MNIST dataset contains $70000$ handwritten images of the digits 0 to 9~\cite{lecun1998gradient} and consists of images that were originally in the National Institute of Standards and Technology (NIST) dataset. The images are grayscale with the digits centered in the $28\times 28$ images.

\subsection{SEALS dataset}
\begin{figure}[tbp]
\centering
\includegraphics[width=0.6\columnwidth, keepaspectratio]{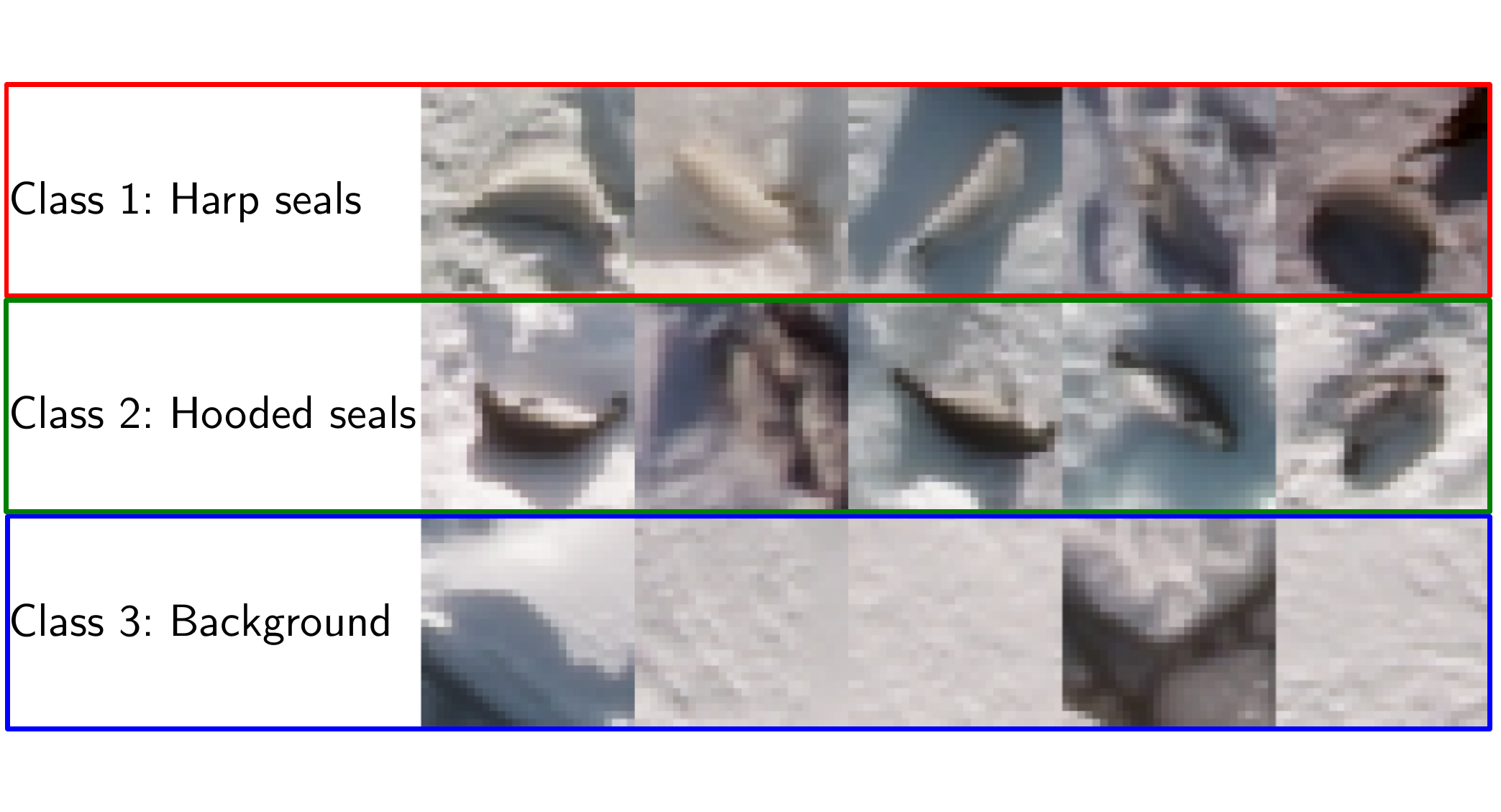}
\caption{Examples from the SEALS-3 dataset. The top row displays examples from the harp seal class, the middle row from the hooded seal, and the bottom row from the background class.}
\label{fig:seals}
\end{figure}

The SEALS-3 dataset consists of several thousand aerial RGB images acquired during surveys in the West Ice east of Greenland in 2007 and 2012 and east of New Foundland, Canada, in 2012. The images are acquired from approximately 300m altitude, and the pixel spacing is about 3cm (depending on the exact flight altitude). A typical image size is $11500\times7500$ pixels. From these images $64\times64$ image crops of harp seal pups, hooded seal pups and background (non-seals) were extracted and down-sampled to $28\times28$ to fit our chosen architecture.
As the smallest class consists of $1000$ images, we select a reduced set of $1000$ images from each class to create a balanced dataset.
Figure~\ref{fig:seals} shows example images for the three classes in the SEALS-3 dataset.

As the background class contains a large variety of images, such as white snow and black water images, unsupervised algorithms are likely to partition these instances into different clusters. Therefore, we additionally created and tested another dataset (SEALS-2), where the background class was not included.

\subsection{Reuters dataset}
The Reuters dataset consists of $800000$ news stories that have been manually categorized into a category tree~\cite{lewis2004rcv1}. In this work, similarly to~\cite{xie2015unsupervised}, we chose the four root categories as labels, removed stories that are labeled with multiple root categories and represent each news story as a feature vector consisting of the Term Frequency-Inverse Document Frequency (TF-IDF) of the 2000 most frequently occurring word stems. As done for the SEALS dataset, we select $54000$ datapoints from each class in order to balance the dataset.

\subsection{Performance measures}
To evaluate the partition quality obtained after training, we consider two different supervised measures.
The first measure is the Normalized mutual information (NMI), defined as
\begin{equation}
\label{eq:nmi}
NMI = \frac{I(l,c)}{\frac{1}{2}[H(l)+H(c)]} \; ,
\end{equation}
where $I(\cdot,\cdot)$ and $H(\cdot)$ denote mutual information and entropy functionals, respectively.
The second evaluation metric is the unsupervised clustering accuracy
\begin{equation}
\label{eq:acc}
ACC = \max_\mathcal{M} \frac{\sum_{i=1}^{n}\delta(l_i=\mathcal{M}(c_i))}{n} \; ,
\end{equation}
where $l_i$ refers to the ground truth label, $c_i$ to the assigned cluster of data point $i$, and $\delta(\cdot)$ is the Dirac delta. $\mathcal{M}$
is the mapping function that corresponds to the optimal one-to-one assignment of clusters to label classes implemented by means of the Hungarian
algorithm~\cite{kuhn1955hungarian}, which solves the linear assignment problem of assigning a cluster to its label class in polynomial time.

\subsection{Baseline methods}
As methods for comparison, we use $k$-means (with the so-called $k$-means++ initialization \cite{Arthur:2007:KAC:1283383.1283494}) as a well-known baseline and a hierarchical information theoretic clustering approach~\cite{Vikjord20143070} based on implicit cluster density estimation using 1) a k-NN approach (ITC-kNN) and 2) a parzen window approach (ITC-parzen).
Further we compare our approach to a representative subset of state-of-the-art methods in clustering, namely \textit{Deep Embedded Clustering} (DEC) \cite{xie2015unsupervised}\footnote{Caffe version of DEC publicly available: \url{https://github.com/piiswrong/dec}}, \textit{Spectral Embedded Clustering} (SEC) \cite{nie2011spectral}, and \textit{Local Discriminant Models and Global Integration} (LDMGI) \cite{yang2010image}.
SEC and LDMGI are spectral clustering algorithms based on the foundations discussed in \cite{ng2002spectral}.
In SEC, the authors jointly optimize the normalized cut loss function and a linear transformation from input to the embedding space for spectral clustering, such that the transformed data is close to the embedded data.
The similarity is modeled using a Gaussian kernel.
LDMGI optimizes a similar objective function, but the Laplacian matrix is learned by exploiting manifold structure and discriminant information, in contrast to most spectral clustering methods where the Laplacian is calculated by using a Gaussian kernel.
Both of these methods use \textit{spectral rotation} \cite{yu2003multiclass} to obtain the final cluster assignments instead of $k$-means, which is common for many spectral clustering methods.
To the authors best knowledge, these two methods represent the current state-of-the-art in spectral clustering, outperforming conventional spectral clustering methods in a wide variety of clustering tasks~\cite{nie2011spectral, yang2010image}.

Following the experiment setup of~\cite{xie2015unsupervised}, the parameters in the baseline models are set according to the suggestions in their respective papers, varying their hyperparameters over 9 possible choices. 
For each one, we run the baseline models 20 times. The best result is shown in the experiments. Due to the lack of hyperparameters in $k$-means (except the number of clusters $k$, which is fixed in our experiments), the accuracy for the best run from 20 different random initializations is reported.

\subsection{Implementation}
\label{sec:Implementation}
The proposed network model is trained end-to-end by using Adam~\cite{kingma2014adam} and implemented using the Theano framework~\cite{theano}. For each image datasets we used the same convolutional architecture and for each vectorial datasets we used the same fully connected architecture. Training is performed on mini-batches of size $100$.
By avoiding a fine-tuning for each problem at hand, we show the robustness of our architecture.
Training is performed with a learning rate of $0.001$ for the convolutional neural network and $10^{-5}$ for the fully connected network. The network is trained for 70000 iterations and the ordering of the mini-batches was reshuffled at each epoch. Weights of the network are initialized following~\cite{he2015delving}. Following the rule-of-thumb in~\cite{jenssenKECA}, $\sigma$ of the Gaussian kernel was chosen to be $15\%$ of the median pairwise Euclidean distances between the feature representation produced by the first fully-connected layer, which produced satisfying results for all datasets. 
The median is adaptive and recomputed as part of the cost function evaluation. In our experiments, we observed that this rule-of-thumb benefited considerably from activation rescaling through batch-normalization.

As \dc{} is prone to get stuck in local minima, a common problem for unsupervised deep architectures, we run \dc{} for $20$ runs and report the accuracy of the run with the lowest value of our unsupervised loss function.
We also report the results of a voting scheme of the top three runs according to our unsupervised loss function. Following~\cite{strehl2002cluster, Vikjord20143070}, clustering results of the best performing run are used as a starting point and the clustering results of the other two runs are aligned to it via the mapping function provided by the Hungarian algorithm in an unsupervised manner. Once the results are aligned, we combine them via a simple voting procedure and compute the final unsupervised clustering accuracy using \eqref{eq:acc}. In the following, this network ensemble is referred to as \dc{}-VOTE. Note that the voting procedure is completely unsupervised and is commonly used in ensemble approaches.

\subsection{Results}

\begin{figure*}[tbp]
\centering
\includegraphics[width=0.7\textwidth,keepaspectratio]{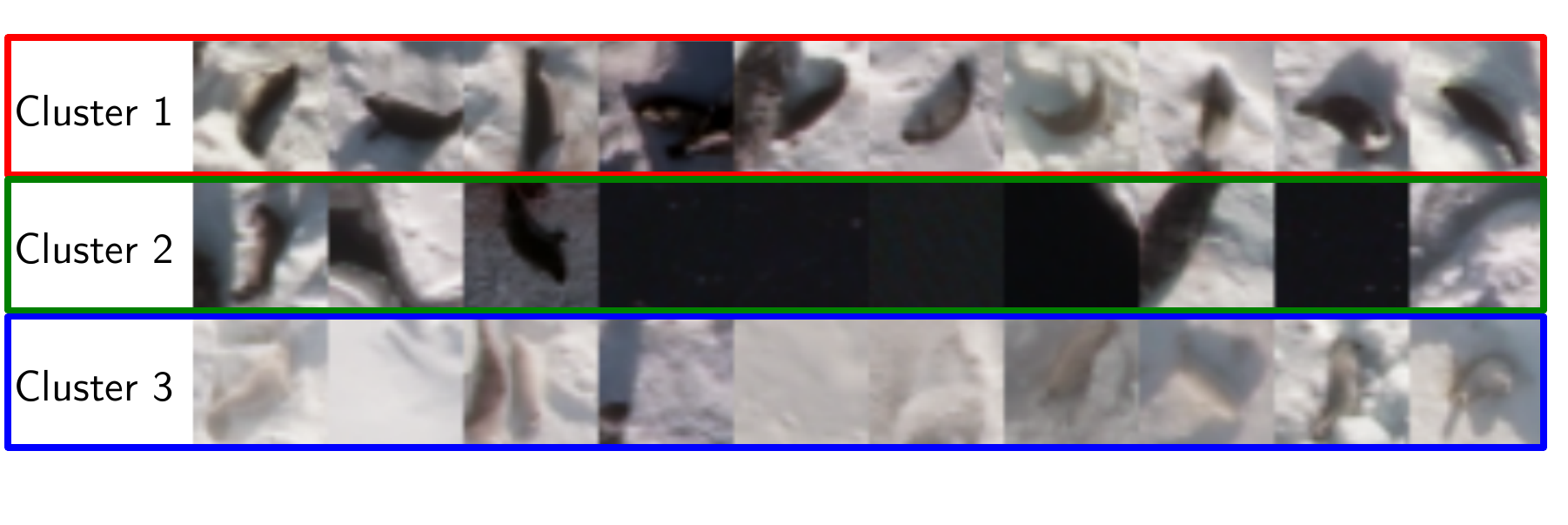}
\caption{Clustering result for the three classes in the SEALS-3 dataset. The first cluster appears to correspond to hooded seals, whereas the other two clusters correspond to a mix of background and seal images.}\label{fig:res_3_class}
\end{figure*}

\begin{figure*}[tbp]
\centering
\begin{subfigure}[b]{0.48\textwidth}
  \centering
  \includegraphics[width=\textwidth,keepaspectratio,trim={1.18cm 0.76cm 0.22cm 0.4cm},clip]{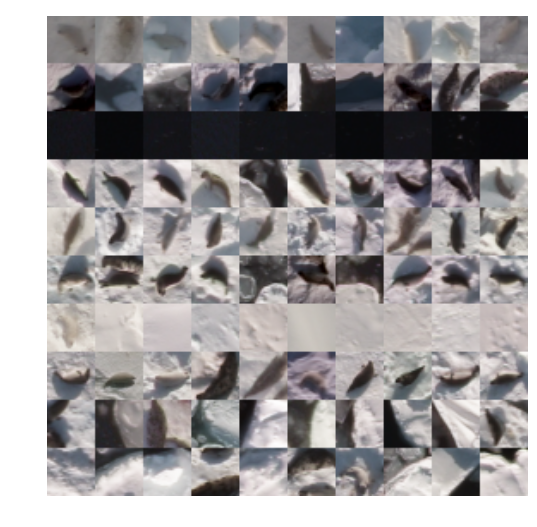}
  \caption{}
  \label{fig:seals_10_class}
\end{subfigure}\quad%
\begin{subfigure}[b]{0.48\textwidth}
  \centering
  \includegraphics[width=\textwidth,keepaspectratio,trim={1.19cm 0.85cm 0.22cm 0.42cm},clip]{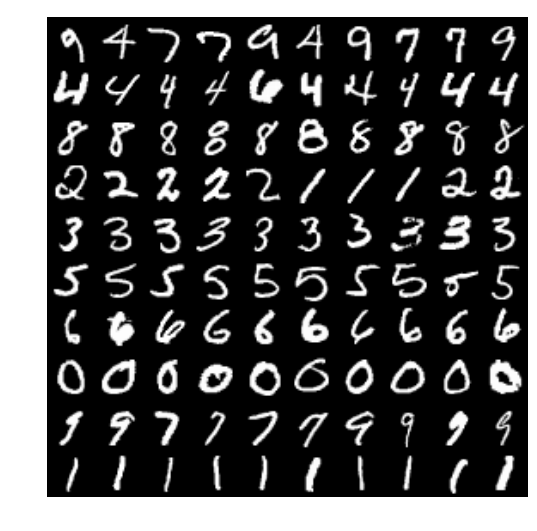}
  \caption{}
  \label{fig:mnist_10_class}
\end{subfigure}
\caption{(a) Results for the SEALS-3 dataset when clustering into ten clusters. (b) Clustering result for the ten-class MNIST dataset.}
\label{fig:res_10_class}
\end{figure*}
We compare \dc{} and \dc{}-VOTE to the baseline algorithms on the MNIST and SEALS datasets and observe that they outperform the baseline methods on all datasets. 
The results can be found in Table \ref{res-table}. Due to very high computational complexity, the ITC algorithm could not be evaluated on MNIST and large datasets in general. This also highlights an important advantage of our formulation with regards to previous clustering approaches based on the CS divergence.
Unlike cluster algorithms that estimate the optimal number of clusters from data, our method and the baseline approaches require the user to specify the number of clusters beforehand. By following a common practice, we have chosen the number of clusters equal to the number of classes in the corresponding datasets. It can be observed that \dc{}-VOTE generally outperforms \dc{}, except in the case of the SEALS-3 dataset, where all tested clustering algorithms perform poorly due to the high variation characterizing the background class. 
Methods based on adversarial networks, namely AAE and CatGAN, have shown to perform well on the MNIST clustering task (the only real dataset analyzed in these papers), by clustering the dataset into a large number of groups ($\ge16$), and mapping these into the $10$ original classes in a post-processing step. 
A similar approach could potentially be employed by \dc{} to boost performance and will be explored in future work.

In what follows, we qualitatively analyze obtained clustering performance. Figure \ref{fig:res_3_class} displays clustering results for the SEALS-3 dataset, where each row corresponds to the top ten scoring images for each of the three clusters.
It is possible to note that the clustering result for the second and third cluster corresponds to a mix of background and seal images, with the third one containing white background images and harp seals and the second cluster containing black background images and hooded seals. As clustering is an unsupervised task and does not necessarily agree with the available supervised information, this result is not unexpected due to the fact that the background class includes large variations.

To further illustrate the clusters found in the dataset, we increased the number of clusters for the SEALS-3 dataset to $10$. Figure \ref{fig:seals_10_class} shows an overview over the learned clusters. It can be observed that \dc{} generally finds reasonable clusters, for example by grouping water (dark patches) in one cluster and white background images in another. Also, it is possible to note that \dc{} generally groups the two different seal classes into separate clusters and assigns images containing both water and snow to a specific class. 

\begin{figure}[tbp]
\centering
\includegraphics[width=\columnwidth,keepaspectratio]{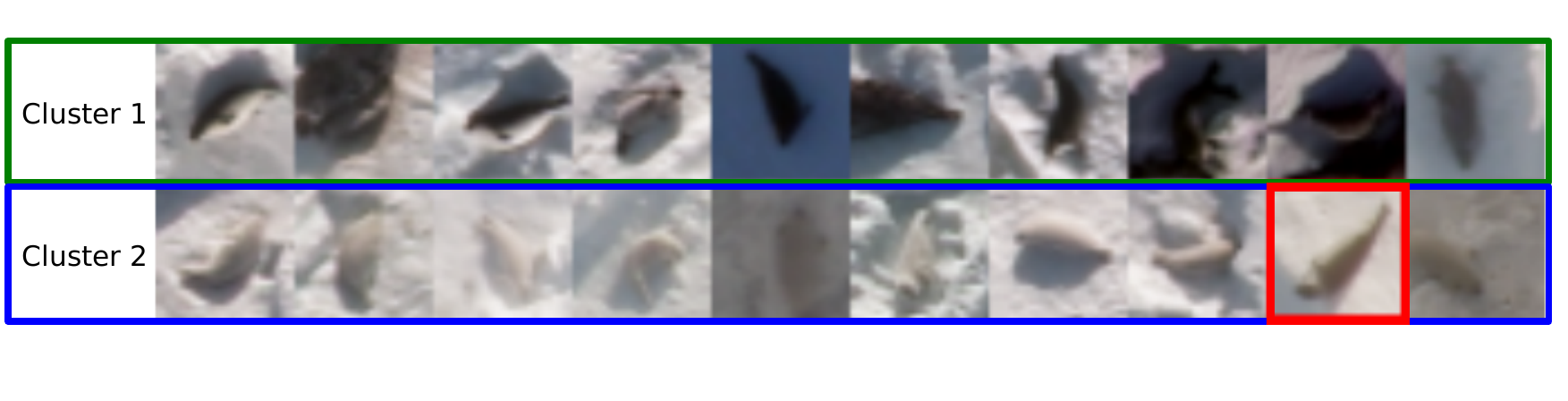}
\caption{Clustering result for the two classes in the SEALS-2 dataset. \dc{} groups the two seal types in distinct clusters. The dark seals in the top row corresponding to hooded seals and the light seals in the bottom row corresponding to harp seals. The red box indicates a mismatch.}\label{fig:res_2_class}
\end{figure}
\bgroup
\def\arraystretch{1}
\setlength\tabcolsep{.6em}
\begin{table}[tbp] \small
\centering
\caption{Clustering accuracy for \dc{}, \dc{}-VOTE, and the baseline models. Best results are highlighted in bold.}
\label{res-table}
\begin{tabular}{l|ccc}
\cmidrule[1.5pt]{1-4}
\multicolumn{1}{l}{\bf Datasets} & {\bf Method}& {\bf NMI} & {\bf ACC[\%]}\\
\cmidrule[1.5pt]{1-4}
\multirow{8}{*}{MNIST} & K-means & 0.50 & 53.33 \\
& ITC (parzen) & - & - \\
& ITC (kNN) & - & - \\
& SEC & 0.77 & 68.82 \\
& LDMGI & 0.81 & 83.03 \\
& DEC & 0.81 & 84.31 \\
& \dc{} & 0.83 & 86.58\\
& \dc{}-VOTE & {\bf 0.87} & {\bf 88.49}\\
\hline
\multirow{8}{*}{SEALS-3} & K-means & 0.13 & 51.33 \\
& ITC (parzen) & 0.003 & 35.30 \\
& ITC (kNN) & 0.10 & 53.95 \\ 
& SEC & 0.15 & 49.00 \\
& LDMGI & 0.13 & 50.43 \\
& DEC & {\bf0.17} & 50.33 \\
& \dc{} & 0.14 & {\bf 55.97} \\
& \dc{}-VOTE & 0.13 & 53.30 \\
\hline
\multirow{8}{*}{SEALS-2} & K-means & 0.015 & 56.85 \\
& ITC (parzen) & 0.003 & 51.55 \\
& ITC (kNN) & 0.020 & 57.20 \\
& SEC & 0.021 & 58.15 \\
& LDMGI & 0.018 & 57.85 \\
& DEC & 0.005 & 54.04 \\ 
& \dc{} & 0.15  & 72.05 \\ 
& \dc{}-VOTE & {\bf 0.18} & {\bf 74.65}\\ 
\hline
\multirow{8}{*}{Reuters} & K-means & 0.38 & 60.5 \\
& ITC (parzen) & - & - \\
& ITC (kNN) & - & - \\
& SEC & - & - \\
& LDMGI & - & - \\
& DEC & 0.38 & 63.79 \\
& \dc{} & 0.50 & 73.06 \\
& \dc{}-VOTE & {\bf 0.51} & {\bf 77.62} \\
\cmidrule[1.5pt]{1-4}
\end{tabular}
\end{table}
\egroup 
Figure \ref{fig:mnist_10_class} illustrates clustering results on MNIST. Interestingly, each cluster represents a distinct number. However, it can be observed that the $7$'s and $9$'s are mixed, which is expected due to their shape similarity. Furthermore, the MNIST dataset contains $1$'s that are straight and $1$'s that are rotated. Our results indicate that some of the far leaning $1$'s are clustered together with the $2$ class, which has a similar diagonal line. 

The results for the SEALS-3 dataset coincide with our intuition that the background class is likely to be divided into different classes. For the SEALS-2 dataset, we observe that \dc{} outperforms the competitor algorithms by a large margin (Table~\ref{res-table}). As the SEALS dataset is less structured and more challenging datasets, we observe that methods that operate directly in pixel space (i.e., raw input space) perform poorly, stressing the importance of extracting higher-level features for clustering. Figure~\ref{fig:res_2_class} shows clustering results for \dc{}, where it can be clearly observed that hooded seals and harp seals are separated into two distinct clusters. 

The proposed clustering cost function is not dependent on the convolutional architecture used in the previous two experiments and can also be used for training fully connected neural networks -- which are used when working with vectorial data. For this purpose, we consider the Reuters dataset and substitute convolutional layers with fully connected ones as described in Section~\ref{sec:arch}. Results are shown in Table~\ref{res-table}, where it is possible to observe that \dc{} outperforms the competitors. Note that, due to the size of the Reuters dataset, running LDMGI and SEC was impossible as a consequence of their memory requirements discussed in Section~\ref{sec:memory}. Note that, from the results presented in~\cite{xie2015unsupervised}, we can see that DEC still performs well when handling the imbalanced Reuters dataset, where the balanced assumption of \dc{} does not hold.
\begin{figure*}[tbp]
\centering
\begin{subfigure}[b]{0.48\columnwidth}
  \centering
  \includegraphics[width=\textwidth, keepaspectratio]{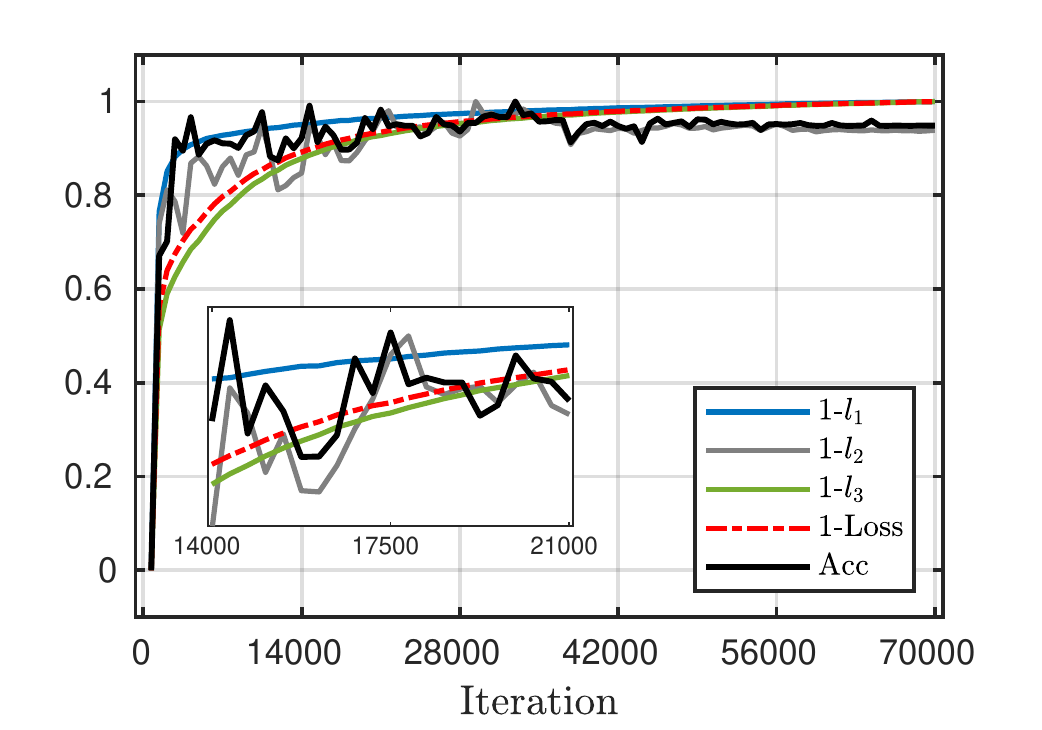}
  \caption{}
  \label{fig:loss}
\end{subfigure}%
\quad
\begin{subfigure}[b]{0.48\columnwidth}
  \centering
  \includegraphics[width=\textwidth, keepaspectratio]{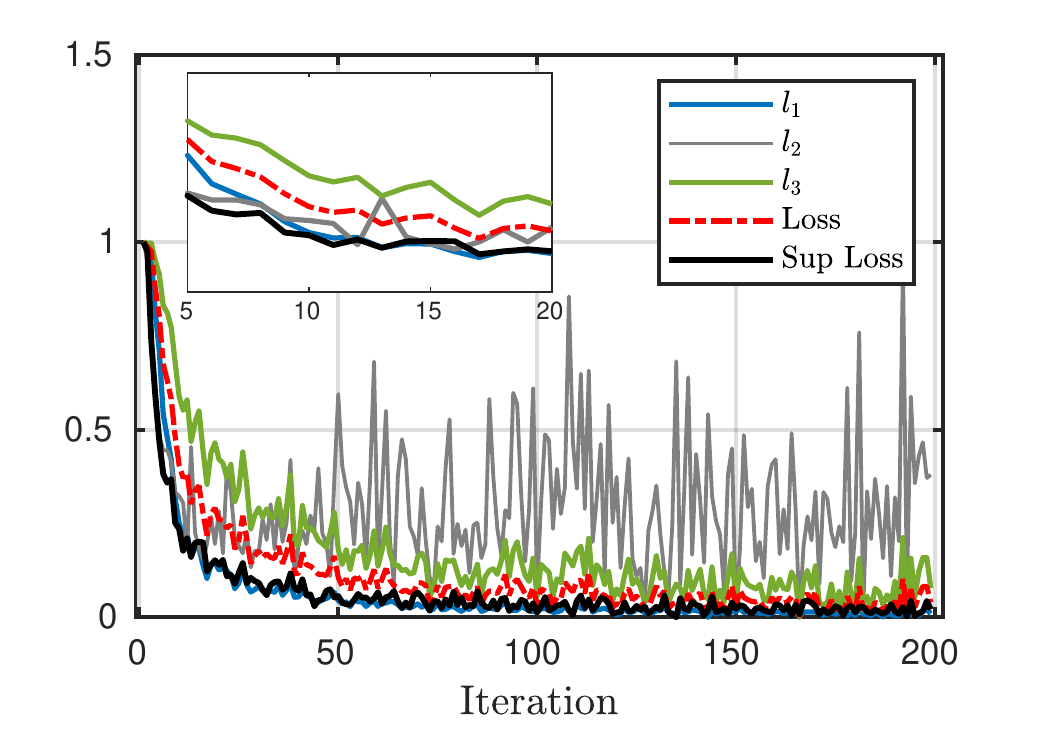}
  \caption{}
  \label{fig:sup_loss}
\end{subfigure}
\caption{(a) Comparison of the supervised accuracy with the unsupervised loss function (Loss) and its three terms ($l_1=d_\text{hid,$\alpha$}$,$l_2=\triu(\mathbf A \mathbf A^T)$, and $l_3=d_\text{hid,m}$) during training. (b) Values of the loss function and its three terms when training the network using a supervised loss function (Sup Loss). Note, the losses have been rescaled to range [0,1].}
\label{fig:costFunc}
\end{figure*}

\subsection{Loss function}
In the following, we analyze the proposed loss function \eqref{eq:loss_fun}, providing empirical evidence of the importance of the different terms; moreover, we evaluate whether the different terms in the loss are related to the performance of the model. Figure \ref{fig:loss} shows the different terms in the loss function during the training phase as we monitor the accuracy of the best run on MNIST.
It is clearly possible to observe that all terms (and also the overall loss) agree reasonably well with the overall clustering accuracy.

\begin{table}[tbp] \small
\def\arraystretch{1} 
\setlength\tabcolsep{.6em}
\centering
\caption{Results of the ablation experiment for the MNIST and the Reuters datasets, illustrating the effect of the three different terms ($l_1=d_\text{hid,$\alpha$}$,$l_2=\triu(\mathbf A \mathbf A^T)$, and $l_3=d_\text{hid,m}$) composing the loss function \eqref{eq:loss_fun}.}

\label{tab:ablation}
\begin{tabular}{l|cc|cc}
\multicolumn{1}{l}{} & \multicolumn{2}{c}{MNIST} & \multicolumn{2}{c}{Reuters}\\
\cmidrule[1.5pt]{1-5}
{\bf Cost} & {\bf Mean $\pm$ std} & {\bf Max} & {\bf Mean $\pm$ std} & {\bf Max}\\
\cmidrule[1.5pt]{1-5}
l1 & $26.1 \pm 2.9$ & $31.5$ & $39.3 \pm 4.8$ & $48.7$\\
l2 & $48.9 \pm 1.5$ & $51.8$ & $41.1 \pm 2.9$ & $46.0$\\
l3 & $74.0 \pm 2.9$ & $77.0$ & $70.2 \pm 6.4$ & $78.8$\\
\hline
l1+l2 & $71.9 \pm 8.0$ & $83.5$  & $66.0 \pm 8.2$ & $75.9$\\
l1+l3 & $84.7 \pm 5.1$ & $88.6$ & $64.8 \pm 5.2$ & $72.7$\\
l2+l3 & $74.8 \pm 5.5$ & $84.3$ & $70.3 \pm 4.4$ & $75.4$ \\
\hline
l1+l2+l3 & $80.8 \pm 5.8$ & $87.5$ & $69.8 \pm 7.3$ & $82.6$ \\
\cmidrule[1.5pt]{1-5}
\end{tabular}

\end{table}

Considering that the network architecture is identical to networks used in supervised approaches and the availability of labels in our datasets, we can also monitor the terms of the loss function during supervised training. 
Figure \ref{fig:sup_loss} shows how each term in the loss function and the overall loss decrease as we perform supervised training on MNIST using a cross-entropy loss function. Again, it is possible to notice that the individual terms have a similar decreasing trend. Note that large variations in the second loss term correspond to the aforementioned regularization effect (see Sect. \ref{sec:regularization}).

In order to further analyze our method, we perform an ablation experiment~\cite{nguyen2015deep} to investigate the effect of the different terms on the clustering result. To this end, we recompute clustering accuracy on MNIST and Reuters for all different combinations of cost function terms. We repeat the experiments five times (20 runs each), compute the overall accuracy for the run with the lowest cost function each time and report the mean, standard deviation and the maximum accuracy values over these five results. The maximum accuracy value is reported solely in order to illustrate the maximum potential of the proposed method. In practice, the strategy from Section~\ref{sec:Implementation} would be used, wherein the best models according to the unsupervised cost function from each run are combined, denoted as DDC-Vote.
Note, this differs from the results reported in Table~\ref{res-table} for DDC, where we report the accuracy only over $20$ runs.
Results for the MNIST and Reuters datasets are reported in Table~\ref{tab:ablation}.
Our results illustrate that, by using all three terms together, we generally obtain better performance. However, the contribution of each term to the final performance is not consistent over all datasets. For instance, we observe that the $l2$ regularization term does not improve the overall result on the MNIST dataset, but does have a positive effect on the Reuters dataset.

\begin{figure}[tbp]
  \centering
  \includegraphics[width=0.4\columnwidth, keepaspectratio,clip]{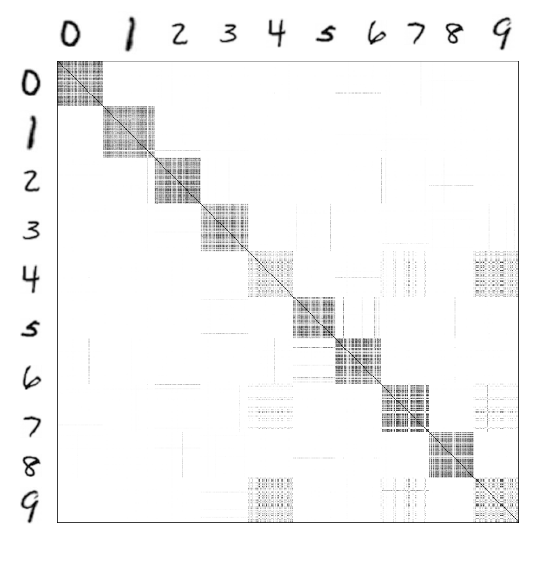}
  \caption{Kernel matrix computed over the learned hidden representation. White colors correspond to low values in the kernel matrix, whereas dark values indicate large values.}
  \label{fig:kernel_mat}
\end{figure}

The three terms in the loss function \eqref{eq:loss_fun} were equally weighted in all experiments. However, better performance might be achieved by weighing such terms according to the data properties. For instance, decreasing the importance on $l_2=\triu(\mathbf A \mathbf A^T)$ might allow our method to better handle imbalanced data sets. The analysis of more advanced weighting schemes is left for future work.

We further conducted experiments where we replaced the CS divergence \eqref{eq:csdiv_parzen} with a symmetrized Kullback-Leibler (KL) divergence, the divergence used by DEC. However, in our experiments we noticed that the performance of the proposed method drops considerably. On MNIST {\dc{}} with KL divergence only obtains an accuracy of $53.66\%$. Further, using a cosine similarity together with the KL divergence for {\dc{}} obtains $35.48\%$ accuracy. We hypothesize that this is mainly due to the fact that the KL divergence encourages separation of clusters, but does not necessarily enforces their compactness. A more thorough analysis of alternative divergence measures is left for future work.

\subsection{Learned representation}
Figure~\ref{fig:kernel_mat} illustrates the final kernel matrix $\mathbf K$ computed over the hidden layer for the best MNIST run. Here, unlike in the case of training, where data points are fed to the model in a random order, the data points have been sorted according to their class labels. A clear block structure is evident from the figure. White and black values indicate low and high similarity, respectively. However, especially the $4$ and $9$ class show high in-between class similarity, which is not surprising due to their closeness in shape.

\subsection{Interpretability of neural network predictions}
\begin{figure}[tbp]
\centering
\minipage[t]{0.1\textwidth}
  \includegraphics[width=\linewidth]{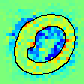}
\endminipage
\minipage[t]{0.1\textwidth}%
  \includegraphics[width=\linewidth]{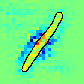}
\endminipage
\minipage[t]{0.1\textwidth}
  \includegraphics[width=\linewidth]{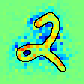}
\endminipage
\minipage[t]{0.1\textwidth}
  \includegraphics[width=\linewidth]{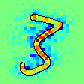}
\endminipage
\minipage[t]{0.1\textwidth}
  \includegraphics[width=\linewidth]{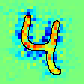}
\endminipage\\
\minipage[t]{0.1\textwidth}%
  \includegraphics[width=\linewidth]{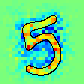}
\endminipage
\minipage[t]{0.1\textwidth}%
  \includegraphics[width=\linewidth]{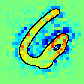}
\endminipage
\minipage[t]{0.1\textwidth}%
  \includegraphics[width=\linewidth]{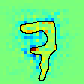}
\endminipage
\minipage[t]{0.1\textwidth}%
  \includegraphics[width=\linewidth]{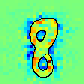}
\endminipage
\minipage[t]{0.1\textwidth}%
  \includegraphics[width=\linewidth]{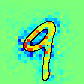}
\endminipage
\caption{Interpretability results for a random subset of MNIST examples using the guided backpropagation technique.}\label{fig:guided_backprop}
\end{figure}

A recent trend in deep learning is the development of methods to interpret predictions of neural networks trained with supervised information~\cite{springenberg2014striving,montavon2017explaining}. However, interpretability is not only a problem in supervised settings.
It could be argued that it is even more essential for unsupervised training, where learning is task-driven. One such approach is the guided backpropagation~\cite{springenberg2014striving}, which allows visualizing pixels in the image that are most influential for a given output decision.
We use this technique to visualize what the model learns to recognize MNIST images. Figure~\ref{fig:guided_backprop} illustrates the results for $10$ random numbers of the MNIST dataset. Red pixels indicate pixels positively correlated with the output, in our case the input to the softmax layer (the unnormalized class score) and lowering the value of the red pixels will lead to a reduced class score. We observe that our method focuses on features that are unique for a specific number. This includes, for instance, the loop for the $6$ and the top part of the $4$.
Rendering of the interpretability results overlaying the MNIST number was inspired by \cite{lrptoolbox}.

\section{Conclusion and future work}
\label{sec:conclusion}
In this paper, we proposed a novel approach to clustering, dubbed \dc{}, which (i) takes advantage of the power of deep learning architectures, (ii) is trainable end-to-end in a fully unsupervised way, (iii) does not require pre-training or complex feature design such as HOG~\cite{dalal2005histograms} and SIFT~\cite{lowe2004distinctive}, and finally (iv) achieves results that outperform or are comparable with state-of-the-art methods on two real-world image datasets and a news story text dataset.
We have also evaluated the performance of an ensemble \dc{} approach, which generally outperformed a single \dc{} model in the considered benchmarks.
Overall, experimental results presented here are promising and stress the importance of unsupervised learning in modern deep learning methods.

In future works, we intend to study the robustness of our method. Further, we will explore alternative loss function formulations, including approaches that are not based on divergence measures and information-theoretic learning.
Finally, it would be interesting to explore the use of the proposed method for related clustering settings, such as for instance multi-view clustering~\cite{nie2018multiview} and constraint clustering~\cite{li2018rank}.

\section*{Acknowledgment}
We gratefully acknowledge the support of NVIDIA Corporation with the donation of the GPU used for this research. This work was partially funded by the Norwegian Research Council FRIPRO grant no.\ 239844 on developing the \emph{Next Generation Learning Machines} and IKTPLUSS grant no. 270738 \emph{Deep Learning for Health}.
Thanks to the Research Council of Norway (UAVSEAL project, no. 234339) for funding, and the Institute of Marine Research, Norway, and Northwest Atlantic Fisheries Centre, Canada, for providing images and ground truth information about the location of the seal pups in the images.


\section*{References}
\bibliographystyle{elsarticle-num}
\bibliography{ref.bib}

\begin{thebibliography}{10}
\expandafter\ifx\csname url\endcsname\relax
  \def\url#1{\texttt{#1}}\fi
\expandafter\ifx\csname urlprefix\endcsname\relax\def\urlprefix{URL }\fi
\expandafter\ifx\csname href\endcsname\relax
  \def\href#1#2{#2} \def\path#1{#1}\fi

\bibitem{krizhevsky2012imagenet}
A.~Krizhevsky, I.~Sutskever, G.~E. Hinton, Imagenet classification with deep
  convolutional neural networks, in: Advances in neural information processing
  systems, 2012, pp. 1097--1105.

\bibitem{Goodfellow-et-al-2016}
I.~Goodfellow, Y.~Bengio, A.~Courville, Deep Learning, MIT Press, 2016,
  \url{http://www.deeplearningbook.org}.

\bibitem{bengio2013representation}
Y.~Bengio, A.~Courville, P.~Vincent, Representation learning: {A} review and
  new perspectives, IEEE Transactions on Pattern Analysis and Machine
  Intelligence 35~(8) (2013) 1798--1828.
\newblock \href {http://dx.doi.org/10.1109/TPAMI.2013.50}
  {\path{doi:10.1109/TPAMI.2013.50}}.

\bibitem{renNIPS15fasterrcnn}
S.~Ren, K.~He, R.~Girshick, J.~Sun, Faster {R-CNN}: Towards real-time object
  detection with region proposal networks, in: Advances in Neural Information
  Processing Systems ({NIPS}), 2015.

\bibitem{long2015fully}
J.~Long, E.~Shelhamer, T.~Darrell, Fully convolutional networks for semantic
  segmentation, in: Proceedings of the IEEE Conference on Computer Vision and
  Pattern Recognition, 2015, pp. 3431--3440.

\bibitem{kampffmeyer2016semantic}
M.~Kampffmeyer, A.-B. Salberg, R.~Jenssen, Semantic segmentation of small
  objects and modeling of uncertainty in urban remote sensing images using deep
  convolutional neural networks, in: Proceedings of the IEEE Conference on
  Computer Vision and Pattern Recognition Workshops, 2016, pp. 1--9.

\bibitem{bianchi2017overview}
F.~M. Bianchi, E.~Maiorino, M.~C. Kampffmeyer, A.~Rizzi, R.~Jenssen, An
  overview and comparative analysis of recurrent neural networks for short term
  load forecasting, arXiv preprint arXiv:1705.04378.

\bibitem{graves2013speech}
A.~Graves, A.-r. Mohamed, G.~Hinton, Speech recognition with deep recurrent
  neural networks, in: 2013 IEEE international conference on acoustics, speech
  and signal processing, IEEE, 2013, pp. 6645--6649.

\bibitem{jain2010data}
A.~K. Jain, Data clustering: 50 years beyond k-means, Pattern Recognition
  Letters 31~(8) (2010) 651--666.

\bibitem{von2007tutorial}
U.~Von~Luxburg, A tutorial on spectral clustering, Statistics and computing
  17~(4) (2007) 395--416.

\bibitem{xie2015unsupervised}
J.~Xie, R.~Girshick, A.~Farhadi, Unsupervised deep embedding for clustering
  analysis, in: Proceedings of the 33rd International Conference on
  International Conference on Machine Learning, Vol.~48, JMLR.org, 2016, pp.
  478--487.

\bibitem{yang2016towards}
B.~Yang, X.~Fu, N.~D. Sidiropoulos, M.~Hong, Towards k-means-friendly spaces:
  Simultaneous deep learning and clustering, arXiv preprint arXiv:1610.04794.

\bibitem{goodfellow2014generative}
I.~Goodfellow, J.~Pouget-Abadie, M.~Mirza, B.~Xu, D.~Warde-Farley, S.~Ozair,
  A.~Courville, Y.~Bengio, Generative adversarial nets, in: Advances in Neural
  Information Processing Systems, 2014, pp. 2672--2680.

\bibitem{springenberg2015unsupervised}
J.~T. Springenberg, Unsupervised and semi-supervised learning with categorical
  generative adversarial networks, arXiv preprint arXiv:1511.06390.

\bibitem{makhzani2015adversarial}
A.~Makhzani, J.~Shlens, N.~Jaitly, I.~Goodfellow, B.~Frey, Adversarial
  autoencoders, arXiv preprint arXiv:1511.05644.

\bibitem{dhillon2003divisive}
I.~S. Dhillon, S.~Mallela, R.~Kumar, A divisive information-theoretic feature
  clustering algorithm for text classification, Journal of machine learning
  research 3~(Mar) (2003) 1265--1287.

\bibitem{jenssen2007information}
R.~Jenssen, D.~Erdogmus, K.~E. Hild, J.~C. Principe, T.~Eltoft, Information cut
  for clustering using a gradient descent approach, Pattern Recognition 40~(3)
  (2007) 796--806.

\bibitem{tishby2001data}
N.~Tishby, N.~Slonim, {Data clustering by Markovian relaxation and the
  Information Bottleneck Method}, in: Advances in Neural Information Processing
  Systems, Vol.~{13}, {2001}, pp. {640--646}.

\bibitem{Vikjord20143070}
V.~V. Vikjord, R.~Jenssen, Information theoretic clustering using a k-nearest
  neighbors approach, Pattern Recognition 47~(9) (2014) 3070--3081.
\newblock \href {http://dx.doi.org/10.1016/j.patcog.2014.03.018}
  {\path{doi:10.1016/j.patcog.2014.03.018}}.

\bibitem{aragam2018identifiability}
B.~Aragam, C.~Dan, P.~Ravikumar, E.~P. Xing, Identifiability of nonparametric
  mixture models and {B}ayes optimal clustering, arXiv preprint
  arXiv:1802.04397.

\bibitem{kampffmeyer2017mlsp}
M.~Kampffmeyer, S.~L{\o}kse, F.~M. Bianchi, L.~Livi, A.-B. Salberg, R.~Jenssen,
  Deep divergence-based clustering, in: Proceedings of the IEEE Workshop on
  Machine Learning for Signal Processing, Tokyo, Japan, 2017.

\bibitem{springenberg2014striving}
J.~T. Springenberg, A.~Dosovitskiy, T.~Brox, M.~Riedmiller, Striving for
  simplicity: The all convolutional net, arXiv preprint arXiv:1412.6806.

\bibitem{erhan2010does}
D.~Erhan, Y.~Bengio, A.~Courville, P.-A. Manzagol, P.~Vincent, S.~Bengio, Why
  does unsupervised pre-training help deep learning?, Journal of Machine
  Learning Research 11~(Feb) (2010) 625--660.

\bibitem{hinton2006fast}
G.~E. Hinton, S.~Osindero, Y.-W. Teh, A fast learning algorithm for deep belief
  nets, Neural Computation 18~(7) (2006) 1527--1554.

\bibitem{vincent2008extracting}
P.~Vincent, H.~Larochelle, Y.~Bengio, P.-A. Manzagol, Extracting and composing
  robust features with denoising autoencoders, in: Proceedings of the 25th
  international conference on Machine learning, ACM, 2008, pp. 1096--1103.

\bibitem{rifai2011contractive}
S.~Rifai, P.~Vincent, X.~Muller, X.~Glorot, Y.~Bengio, Contractive
  auto-encoders: Explicit invariance during feature extraction, in: Proceedings
  of the 28th international conference on machine learning (ICML-11), 2011, pp.
  833--840.

\bibitem{Kampffmeyer2017}
M.~Kampffmeyer, S.~L{\o}kse, F.~M. Bianchi, R.~Jenssen, L.~Livi, Deep
  kernelized autoencoders, in: Scandinavian Conference on Image Analysis,
  Springer, 2017, pp. 419--430.

\bibitem{kingma2013auto}
D.~P. Kingma, M.~Welling, Auto-encoding variational bayes, arXiv preprint
  arXiv:1312.6114.

\bibitem{gregor2015draw}
K.~Gregor, I.~Danihelka, A.~Graves, D.~J. Rezende, D.~Wierstra, Draw: A
  recurrent neural network for image generation, arXiv preprint
  arXiv:1502.04623.

\bibitem{sohn2015learning}
K.~Sohn, H.~Lee, X.~Yan, Learning structured output representation using deep
  conditional generative models, in: Advances in Neural Information Processing
  Systems, 2015, pp. 3483--3491.

\bibitem{rodriguez2014clustering}
A.~Rodriguez, A.~Laio, Clustering by fast search and find of density peaks,
  Science 344~(6191) (2014) 1492--1496.

\bibitem{cagata}
F.~M. Bianchi, L.~Livi, A.~Rizzi, Two density-based k-means initialization
  algorithms for non-metric data clustering, Pattern Analysis and Applications
  19~(3) (2016) 745--763.
\newblock \href {http://dx.doi.org/10.1007/s10044-014-0440-4}
  {\path{doi:10.1007/s10044-014-0440-4}}.

\bibitem{nie2018multiview}
F.~Nie, L.~Tian, X.~Li, Multiview clustering via adaptively weighted
  procrustes, in: Proceedings of the 24th ACM SIGKDD International Conference
  on Knowledge Discovery \& Data Mining, ACM, 2018, pp. 2022--2030.

\bibitem{myhre2018robust}
J.~N. Myhre, K.~{\O}. Mikalsen, S.~L{\o}kse, R.~Jenssen, Robust clustering
  using a knn mode seeking ensemble, Pattern Recognition 76 (2018) 491--505.

\bibitem{comaniciu2002mean}
D.~Comaniciu, P.~Meer, Mean shift: A robust approach toward feature space
  analysis, IEEE Transactions on pattern analysis and machine intelligence
  24~(5) (2002) 603--619.

\bibitem{aggarwal2013data}
C.~C. Aggarwal, C.~K. Reddy, Data Clustering: Algorithms and Applications, CRC
  Press, Boca Raton, Florida, US, 2013.

\bibitem{jenssenKECA}
R.~Jenssen, Kernel entropy component analysis, IEEE Transactions on Pattern
  Analysis and Machine Intelligence 32~(5) (2010) 847--860.
\newblock \href {http://dx.doi.org/10.1109/TPAMI.2009.100}
  {\path{doi:10.1109/TPAMI.2009.100}}.

\bibitem{ng2002spectral}
A.~Y. Ng, M.~I. Jordan, Y.~Weiss, et~al., On spectral clustering: Analysis and
  an algorithm, Advances in neural information processing systems 2 (2002)
  849--856.

\bibitem{nie2011spectral}
F.~Nie, Z.~Zeng, I.~W. Tsang, D.~Xu, C.~Zhang, Spectral embedded clustering: A
  framework for in-sample and out-of-sample spectral clustering, IEEE
  Transactions on Neural Networks 22~(11) (2011) 1796--1808.

\bibitem{yang2010image}
Y.~Yang, D.~Xu, F.~Nie, S.~Yan, Y.~Zhuang, Image clustering using local
  discriminant models and global integration, IEEE Transactions on Image
  Processing 19~(10) (2010) 2761--2773.

\bibitem{dhillon2004kernel}
I.~S. Dhillon, Y.~Guan, B.~Kulis, Kernel k-means: spectral clustering and
  normalized cuts, in: Proceedings of the tenth ACM SIGKDD international
  conference on Knowledge discovery and data mining, ACM, 2004, pp. 551--556.

\bibitem{han2016mini}
Y.~Han, M.~Filippone, Mini-batch spectral clustering, arXiv preprint
  arXiv:1607.02024.

\bibitem{yan2009fast}
D.~Yan, L.~Huang, M.~I. Jordan, Fast approximate spectral clustering, in:
  Proceedings of the 15th ACM SIGKDD international conference on Knowledge
  discovery and data mining, ACM, 2009, pp. 907--916.

\bibitem{bojanowski2017unsupervised}
P.~Bojanowski, A.~Joulin, Unsupervised learning by predicting noise, arXiv
  preprint arXiv:1704.05310.

\bibitem{yang2016joint}
J.~Yang, D.~Parikh, D.~Batra, Joint unsupervised learning of deep
  representations and image clusters, in: Proceedings of the IEEE Conference on
  Computer Vision and Pattern Recognition, 2016, pp. 5147--5156.

\bibitem{zhang2018multilayer}
X.-L. Zhang, Multilayer bootstrap networks, Neural Networks 103 (2018) 29--43.

\bibitem{dalal2005histograms}
N.~Dalal, B.~Triggs, Histograms of oriented gradients for human detection, in:
  Computer Vision and Pattern Recognition, 2005. CVPR 2005. IEEE Computer
  Society Conference on, Vol.~1, IEEE, 2005, pp. 886--893.

\bibitem{lee2015deeply}
C.-Y. Lee, S.~Xie, P.~Gallagher, Z.~Zhang, Z.~Tu, Deeply-supervised nets., in:
  AISTATS, Vol.~2, 2015, p.~6.

\bibitem{Basseville2013621}
M.~Basseville, Divergence measures for statistical data processing--an
  annotated bibliography, Signal Processing 93~(4) (2013) 621 -- 633.
\newblock \href {http://dx.doi.org/10.1016/j.sigpro.2012.09.003}
  {\path{doi:10.1016/j.sigpro.2012.09.003}}.

\bibitem{jenssen2006cauchy}
R.~Jenssen, J.~C. Principe, D.~Erdogmus, T.~Eltoft, The {C}auchy--{S}chwarz
  divergence and {P}arzen windowing: Connections to graph theory and {M}ercer
  kernels, Journal of the Franklin Institute 343~(6) (2006) 614--629.

\bibitem{gretton2012kernel}
A.~Gretton, K.~M. Borgwardt, M.~J. Rasch, B.~Sch{\"o}lkopf, A.~Smola, A kernel
  two-sample test, Journal of Machine Learning Research 13~(Mar) (2012)
  723--773.

\bibitem{kingma2014adam}
D.~Kingma, J.~Ba, Adam: A method for stochastic optimization, arXiv preprint
  arXiv:1412.6980.

\bibitem{lecun1998gradient}
Y.~LeCun, L.~Bottou, Y.~Bengio, P.~Haffner, Gradient-based learning applied to
  document recognition, Proceedings of the IEEE 86~(11) (1998) 2278--2324.

\bibitem{ioffe2015batch}
S.~Ioffe, C.~Szegedy, Batch normalization: Accelerating deep network training
  by reducing internal covariate shift, arXiv preprint arXiv:1502.03167.

\bibitem{giryes2016deep}
R.~Giryes, G.~Sapiro, A.~M. Bronstein, Deep neural networks with random
  gaussian weights: A universal classification strategy?, IEEE Transactions on
  Signal Processing 64 (2016) 3444--3457.

\bibitem{lewis2004rcv1}
D.~D. Lewis, Y.~Yang, T.~G. Rose, F.~Li, Rcv1: A new benchmark collection for
  text categorization research, Journal of machine learning research 5~(Apr)
  (2004) 361--397.

\bibitem{kuhn1955hungarian}
H.~W. Kuhn, The hungarian method for the assignment problem, Naval research
  logistics quarterly 2~(1-2) (1955) 83--97.

\bibitem{Arthur:2007:KAC:1283383.1283494}
D.~Arthur, S.~Vassilvitskii, {k-means++: the advantages of careful seeding},
  in: {Proceedings of the eighteenth annual ACM-SIAM symposium on Discrete
  algorithms}, {SODA '07}, Society for Industrial and Applied Mathematics,
  Philadelphia, PA, USA, 2007, pp. 1027--1035.

\bibitem{yu2003multiclass}
S.~X. Yu, J.~Shi, Multiclass spectral clustering, in: Computer Vision, 2003.
  Proceedings. Ninth IEEE International Conference on, IEEE, 2003, pp.
  313--319.

\bibitem{theano}
{Theano Development Team}, \href{http://arxiv.org/abs/1605.02688}{{Theano: A
  {Python} framework for fast computation of mathematical expressions}}, arXiv
  e-prints abs/1605.02688.
\newline\urlprefix\url{http://arxiv.org/abs/1605.02688}

\bibitem{he2015delving}
K.~He, X.~Zhang, S.~Ren, J.~Sun, Delving deep into rectifiers: Surpassing
  human-level performance on imagenet classification, in: Proceedings of the
  IEEE international conference on computer vision, 2015, pp. 1026--1034.

\bibitem{strehl2002cluster}
A.~Strehl, J.~Ghosh, Cluster ensembles---a knowledge reuse framework for
  combining multiple partitions, Journal of machine learning research 3~(Dec)
  (2002) 583--617.

\bibitem{nguyen2015deep}
A.~Nguyen, J.~Yosinski, J.~Clune, Deep neural networks are easily fooled: High
  confidence predictions for unrecognizable images, in: Proceedings of the IEEE
  Conference on Computer Vision and Pattern Recognition, 2015, pp. 427--436.

\bibitem{montavon2017explaining}
G.~Montavon, S.~Lapuschkin, A.~Binder, W.~Samek, K.-R. M{\"u}ller, Explaining
  nonlinear classification decisions with deep taylor decomposition, Pattern
  Recognition 65 (2017) 211--222.

\bibitem{lrptoolbox}
S.~Lapuschkin, A.~Binder, G.~Montavon, K.-R. M{{{\"u}}}ller, W.~Samek,
  \href{http://jmlr.org/papers/v17/15-618.html}{The lrp toolbox for artificial
  neural networks}, Journal of Machine Learning Research 17~(114) (2016) 1--5.
\newline\urlprefix\url{http://jmlr.org/papers/v17/15-618.html}

\bibitem{lowe2004distinctive}
D.~G. Lowe, Distinctive image features from scale-invariant keypoints,
  International journal of computer vision 60~(2) (2004) 91--110.

\bibitem{li2018rank}
Z.~Li, F.~Nie, X.~Chang, L.~Nie, H.~Zhang, Y.~Yang, Rank-constrained spectral
  clustering with flexible embedding, IEEE Transactions on Neural Networks and
  Learning Systems.

\end{thebibliography}

\end{document}